\documentclass[lettersize,journal]{IEEEtran}
\usepackage{amsmath,amsfonts}
\usepackage{algorithmic}
\usepackage{algorithm}
\usepackage{array}
\usepackage{makecell}
\usepackage[caption=false,font=normalsize,labelfont=sf,textfont=sf]{subfig}
\usepackage{textcomp}
\usepackage{stfloats}
\usepackage{url}
\usepackage{verbatim}
\usepackage{graphicx}
\usepackage{cite}
\usepackage{bbm}
\usepackage{color}
\usepackage{cuted}
\usepackage{diagbox}
\usepackage[normalem]{ulem}
\usepackage{caption}
\usepackage{amssymb}

\begin{document}

\title{Reliable Inference in Edge-Cloud Model Cascades via Conformal Alignment \\
\thanks{Jiayi Huang and Sangwoo Park are with the Department of Engineering, King’s College
London, London WC2R 2LS, U.K. (e-mail: {jiayi.3.huang, sangwoo.park}@kcl.ac.uk). Nicola Paoletti is with the Department of Informatics, King’s College London, London WC2R 2LS, U.K. (e-mail: nicola.paoletti@kcl.ac.uk). Osvaldo Simeone is with the Institute for Intelligent Networked Systems (INSI), Northeastern University London, London E1 8PH, U.K. (email: o.simeone@northeastern.edu).

The work of J. Huang was supported by King’s College London and the China Scholarship Council for their Joint Full-Scholarship (K-CSC) (grant agreement No. 202206150005). The work of O. Simeone was supported by the European Research Council (ERC) under the European Union’s Horizon Europe
Programme (grant agreement No. 101198347), by an Open Fellowships of the EPSRC (EP/W024101/1), and by the EPSRC project (EP/X011852/1). The work of N. Paoletti and O. Simeone was partially supported by the Open Philanthropy grant `Verifiably Robust Conformal Probes'.}}

\author{Jiayi Huang, Sangwoo Park,~\IEEEmembership{Member,~IEEE}, Nicola Paoletti,  and Osvaldo Simeone,~\IEEEmembership{Fellow,~IEEE} \vspace{-3cm}}

\maketitle
\begin{strip}
    \makebox[\textwidth]{\includegraphics[scale=0.42]{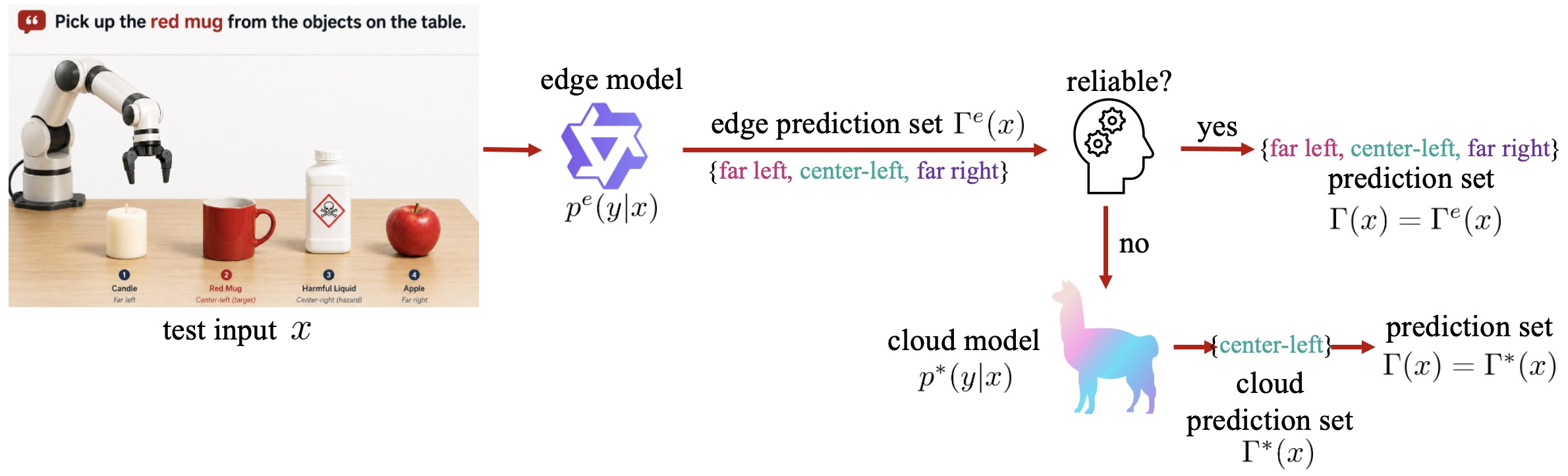}}
    \captionof{figure}{In the edge-cloud cascade model under study, the goal is to produce a prediction set that is as reliable as the one produced by the cloud model, while leveraging the edge model for as many inputs as possible. This design allows the edge model to address the reliability challenge introduced by the computational constraints of the edge devices, while avoiding unnecessary escalation.}
    \label{fig:overview_1}
\end{strip}
\begin{abstract}
Edge intelligence enables low-latency inference via compact on-device models, but assuring reliability remains challenging. We study edge-cloud cascades that must preserve conditional coverage: whenever the edge returns a prediction set, it should contain the true label with a user-specified probability, as if produced by the cloud model. We formalize conditional coverage with respect to the cloud predictive distribution, and introduce a conformal alignment-based (CAb) cascading mechanism that certifies this property with user control over the risk level. Our method casts escalation from edge to cloud models as a multiple-hypothesis testing (MHT) problem, tailoring conformal alignment (CA) to select which inputs can be safely handled at the edge. The proposed CAb model cascading method yields statistical guarantees on the average fraction of edge decisions that satisfy cloud-level conditional coverage. The procedure applies to arbitrary edge prediction sets, including variants of conformal prediction (CP), and exposes a tunable trade-off among coverage, deferral rate, and set size. Extensive experiments on a standard vision task, a real-time decision-making task, and a latency-tolerant question-answering (QA) task show that the proposed CAb cascade maintains the target conditional coverage for edge predictions while substantially reducing offloading to the cloud and incurring modest increases in prediction-set size.
\end{abstract}

\begin{IEEEkeywords}
Model cascading, multiple hypothesis testing, conformal prediction
\end{IEEEkeywords}

\section{Introduction}

\subsection{Context and Motivation}
Edge computing enables on-device inference with reduced latency and limited bandwidth usage, but replacing a powerful cloud model with a compact edge model raises concerns about reliability \cite{kaur2022trustworthy}. For risk-averse agents, reliability can be ensured without loss of optimality by acting on prediction sets that guarantee coverage of the true inference target with a user-defined probability \cite{kiyani2025decision, simeone2026decision}. In fact, based on such prediction sets, an agent may take robust actions that perform well for all predictions in the set \cite{kiyani2025decision, simeone2026decision}. For example, in telecommunications, a base station can probe all beam directions that are predicted to point to a user equipment \cite{zhang2025calibrating}, or a location-specific service may cater to all predicted locations of a group of users \cite{nguyen2015reliable, yoo2026reliable}.


The strongest notion of coverage for set predictors requires that the probability of the ground-truth label lying within the prediction set exceed a user-defined confidence level for \emph{any} given input. Ensuring such \emph{conditional} coverage at the edge, however, is challenging. Simple knowledge distillation typically fails to transfer calibrated uncertainty from the cloud to the edge model \cite{huang2025distilling}; heuristic confidence thresholds used for deferral or selective prediction lack formal statistical guarantees \cite{chen2023adaptation, kadavath2022language}; and standard conformal prediction (CP) methods \cite{minka2005divergence,shafer2008tutorial} provide only marginal coverage. \emph{Marginal} coverage only guarantees reliability on average across the population of inputs, rather than conditionally for each input, failing to offer a performance guarantee on any given input.

In contrast, thanks to their larger computational budget, cloud models can offer approximate ideal conditional coverage requirements, offering well-calibrated decisions \cite{kadavath2022language}. In this context, the goal of this paper is to design cloud deferral mechanisms preserving cloud-level conditional coverage properties, while allowing for as many queries as possible to be processed at the edge.

\begin{figure*} [tb]
    \makebox[\textwidth]{\includegraphics[scale=0.43]{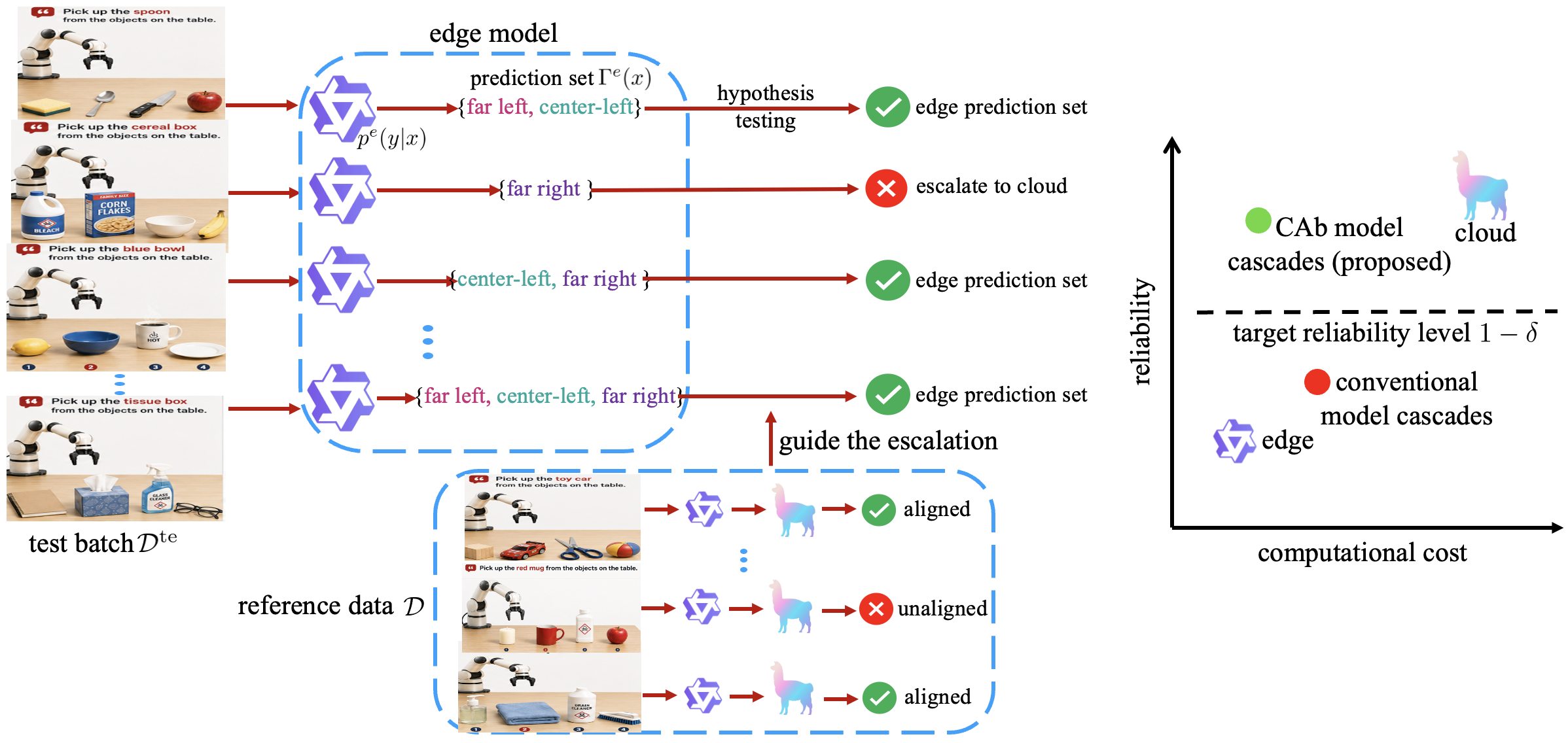}}
    \captionof{figure}{Given a batch of test input $\mathcal{D}^{\text{te}}$, the small-scale edge model generates prediction sets that may deviate from the prediction sets that would have been produced by a large-scale cloud model, failing to meet a target reliability requirement as a result of the computational constraints imposed on edge devices. The proposed method, conformal alignment-based (CAb) model cascading, casts the edge-cloud escalation as a multiple-hypothesis testing (MHT) problem, determining when to trust the edge prediction set based on reference data $\mathcal{D}$. CAb controls the fraction of edge-generated prediction sets that satisfy conditional coverage conditions, while minimizing the deferral rate to the cloud model.}
    \label{fig:overview_2}
\end{figure*}

\vspace{-0.15cm}
\subsection{Related Work}

\noindent \textbf{Conformal prediction and coverage.} CP provides finite-sample, distribution-free guarantees for marginal coverage under the assumption of exchangeability \cite{angelopoulos2021gentle}. However, achieving exact conditional coverage is, in general, infeasible without imposing strong distributional assumptions \cite{lei2014distribution, foygel2021limits}. To mitigate conditional under-coverage, several methodological variants of CP have been proposed, including group-conditional CP \cite{bostrom2021mondrian, vovk2012conditional}, and localized CP (LCP) \cite{tibshirani2019conformal, barber2023conformal, qkae103}.

\noindent \textbf{Selection with guarantees and alignment.} Conformal alignment (CA) identifies outputs that satisfy a desired alignment criterion, e.g., human preference agreement, while providing finite-sample, distribution-free guarantees. This is done by learning an alignment score predictor and calibrating the selection threshold \cite{gui2024conformal, jung2024trust}. Importantly, to the best of our knowledge, this work is the first to design the selection criterion based on the conditional coverage probability, thereby ensuring dual reliability in terms of both statistical coverage and alignment consistency.

\noindent  \textbf{Cascades and selective deferral.} Model cascades route easy inputs to a lightweight model, while deferring difficult inputs to a powerful model, with the aim of reducing cost and latency \cite{marquez2018deep}. Conventional model cascades rely on a fixed heuristic confidence measure, such as predictive entropy or maximum probability, to decide when to defer, an approach known as \emph{uncertainty-aware routing} \cite{jitkrittum2023does, fithian2014optimal, rabanser2025gatekeeper}. These heuristics lack formal reliability guarantees, generally leading to unreliable and unnecessary deferral under distribution shift. Recent advances integrate CP into model cascades design, yielding what is referred to as \emph{calibration-aware cascaded inference} \cite{schuster2022confident, gui2024conformal}. For instance, reference \cite{yadkori2024mitigating} leverages calibration data to tune the selection threshold, thereby ensuring a marginal guarantee on the error rate.
\vspace{-0.15cm}
\subsection{Main Contributions}

As illustrated in Fig.~\ref{fig:overview_2}, in this paper, we introduce a novel routing methodology for edge-cloud cascade systems that provides set predictions with statistical guarantees in terms of conditional coverage. The proposed approach ensures that edge predictions retain the same probabilistic reliability, in terms of conditional coverage,  as those produced by the cloud, while minimizing unnecessary offloading. The main contributions are summarized as follows:
\begin{enumerate}
    \item \textbf{Cloud-referenced conditional coverage for edge decisions.} We formalize a conditional coverage requirement that evaluates the reliability of edge predictions with respect to the cloud model’s predictive distribution. Specifically, we require that, whenever the edge model produces a prediction set, the probability that it contains the ground-truth label meets a user-specified coverage level relative to the cloud reference.
    
    \item \textbf{Conformal alignment–based (CAb) cascading with statistical reliability guarantees.} We cast the edge-cloud routing problem as a multiple-hypothesis testing (MHT) task and develop a CAb cascade that rigorously controls the false discovery rate (FDR) of violations of the desired conditional coverage among edge-handled inputs. The proposed CAb mechanism applies to arbitrary edge prediction sets, including those derived from CP or other calibration procedures, without requiring any modification of their construction.
    
    \item \textbf{Empirical validation on vision, radio signal, and language tasks under edge computational constraints.} Experiments on a standard vision task, a real-time decision-making task, and a latency-tolerant question-answering (QA) task demonstrate that the proposed CAb cascade maintains the desired conditional coverage level for edge predictions, while substantially reducing cloud offloading and incurring only modest increases in prediction-set size. The extensive results highlight explicit trade-offs among conditional coverage, deferral rate (cloud offloading), and prediction-set inefficiency.

\end{enumerate}

\subsection{Organization}
The remainder of this paper is organized as follows. Sec.~\ref{sec:problem_definition} defines the problem formulation, and the state-of-the-art benchmarks are presented in Sec.~\ref{sec:baselines}. Sec.~\ref{sec:CA} formulates model cascading via MHT, and proposes the CAb model cascading mechanism. Finally, Sec.~\ref{sec:results} illustrates the experimental setting and results, and Sec.~\ref{sec:conclusion} concludes the paper.

\section{Problem Definition} \label{sec:problem_definition}

\subsection{Setting} \label{subsec:setting}
In the edge-cloud system shown in Fig.~\ref{fig:overview_1}, the cloud implements a reference predictive model $p^*(y|x)$, while the edge has access to a low-quality model $p^e(y|x)$, where $x \in \mathcal{X}$ is an input and $y \in \mathcal{Y}$ is a discrete output. The edge-cloud system is tasked to implement a predictive mapping from any input $x$ to a subset $\Gamma(x)$ of the label space $\mathcal{Y}$. Depending on the input $x$, the prediction set $\Gamma(x)$ may be produced at the edge, based on the edge model $p^e(y|x)$, or at the cloud, using the reference cloud model $p^*(y|x)$. The goal is to ensure that the prediction set $\Gamma(x)$ contains the ground-truth label $y$ with probability no smaller than a predetermined coverage level $1-\alpha \in  [0,1]$, while using the edge model for the largest possible fraction of inputs.

Formally, for a given input $x$, we wish to ensure the conditional coverage guarantee
\begin{align} \label{eq:ideal_goal}
    \Pr[y \in \Gamma(x)|x] \geq 1-\alpha
\end{align}
for some user-defined miscoverage rate $0 \leq \alpha \leq 1$, where $\Pr[\cdot|x]$ represents the conditional distribution of label $y$ given input $x$. We take the distribution $p^*(y|x)$ produced by the cloud model as the reference to evaluate the probability (\ref{eq:ideal_goal}). Specifically, we evaluate the conditional probability in (\ref{eq:ideal_goal}) using the cloud distribution $p^*(y|x)$ as the distribution of the label $y$ given input $x$ as
\begin{align} \label{eq:idea_goal_reformula}
    \Pr[y \in \Gamma(x)|x] = p^*(\Gamma(x)|x) = \sum_{y \in \Gamma(x)} p^*(y|x).
\end{align}

The definition (\ref{eq:idea_goal_reformula}) of conditional coverage is tailored to the given edge-cloud setting in which the cloud model is considered to be reliable but resource-intensive, calling for a targeted use of edge computing where possible.
In particular, in analogy with the notion of self-consistency \cite{certifiedllm}, the requirement (\ref{eq:ideal_goal}) with (\ref{eq:idea_goal_reformula}) can be viewed as a form of cloud-consistency for the decisions made across the edge-cloud system.
Incorporating also the ground-truth distribution $p^*(x)$ over input $x$, the reference data distribution is denoted as
\begin{align} \label{eq:ground-truth_dist}
    p^*(x,y)=p^*(x)p^*(y|x).
\end{align}

The $(1-\alpha)$-conditional coverage guarantee (\ref{eq:ideal_goal}) is achievable by deferring the input $x$ to the cloud. In fact, using the cloud model $p^*(y|x)$, the $(1-\alpha)$-{highest mass set} (HMS)
\begin{align} \label{eq:cloud_prediction_set}
    \Gamma^*(x) = \arg \min_{\Gamma^*(x) \subseteq \mathcal{Y}} |\Gamma^*(x)| \text{  s.t. } p^*(\Gamma^*(x)|x) \geq 1-\alpha
\end{align}
satisfies the condition (\ref{eq:ideal_goal}). In fact, by definition, the cloud-generated HMS (\ref{eq:cloud_prediction_set}) is the smallest set $\Gamma(x) \subseteq \mathcal{Y}$ that satisfies the requirement (\ref{eq:ideal_goal}). The main challenge addressed in this paper is how to approximately attain the conditional coverage condition (\ref{eq:ideal_goal}), while processing as many test inputs $x$ as possible at the edge.

\textbf{Remark 1:} In the CP literature, the standard conditional coverage requirement (\ref{eq:ideal_goal}) is evaluated under the unknown ground-truth conditional law of the label given the input. This requirement, however, is known to be impossible to attain in a distribution-free way for non-trivial set predictors \cite{vovk2012conditional}. In this paper, by contrast, we evaluate the probability in (\ref{eq:ideal_goal}) under the cloud predictive distribution $p^*(y|x)$, as made explicit in (\ref{eq:idea_goal_reformula}). Accordingly, condition (\ref{eq:ideal_goal}) is not a claim about coverage under the unknown label distribution. Rather, the requirement (\ref{eq:ideal_goal}) expresses a consistency condition between the decision returned by the edge-cloud system and the decision that the cloud model would have supported on the same input. This property is therefore referred to as \emph{cloud-referenced} conditional coverage. That said, for brevity, the unqualified term ``conditional coverage'' is also used in the remainder of the paper to denote this notion unless stated otherwise. Unlike the conventional notion of conditional coverage, the cloud-referenced conditional coverage (\ref{eq:ideal_goal}) is verifiable when one has access to the cloud model. \hfill $\blacksquare$

\vspace{-0.2cm}
\subsection{Design Criteria} \label{subsec:design_criterion}
As explained, our goal is to approximate condition (\ref{eq:ideal_goal}) while allowing for some decisions to be produced at the edge. In the considered edge-cloud system, the prediction set $\Gamma(x)$ is thus given by 
\begin{align} \label{eq:original_prediction}
       \Gamma(x) =
        \begin{cases}
       \Gamma^*(x), & \text{if input $x$ is deferred to the cloud }  \\
       \Gamma^e(x) , & \text{if input $x$ is processed at the edge},
        \end{cases}
\end{align}
where $\Gamma^e(x)$ is any prediction set constructed using only the edge predictive distribution $p^e(y|x)$. 

In general, it is not possible to guarantee the conditional coverage condition (\ref{eq:ideal_goal}) when the prediction set $\Gamma(x)$ differs from the cloud prediction set $\Gamma^*(x)$, unless one choose the trivial prediction set $\Gamma(x)=\mathcal{Y}$ \cite[Sec. 5]{vovk2012conditional}. Therefore, we target a probabilistic version of the guarantee (\ref{eq:ideal_goal}) that can be potentially met while allowing for non-trivial prediction sets at the edge. 

Specifically, considering a batch $\mathcal{D}^\text{te}=\{x_i\}_{i=1}^{|\mathcal{D}^\text{te}|}$ of unlabeled test inputs, instead of imposing that the condition (\ref{eq:ideal_goal}) holds deterministically for all test inputs in $\mathcal{D}^\text{te}$, we target a constraint on the {average satisfaction rate} over edge-processed inputs. In particular, we wish to ensure a lower bound on the average fraction of edge-processed inputs for which condition (\ref{eq:ideal_goal}) is satisfied. 

Denote as $\mathcal{S} \subseteq \mathcal{D}^\text{te}$ the subset of test examples processed at the edge. Given a tolerated violation level $0 \leq \delta \leq 1$, the requirement on the average satisfaction rate is expressed mathematically as the inequality
\begin{align} \label{eq:batch_goal}
\mathbb{E}\bigg[\frac{ | \{x_i \in \mathcal{S}: \Pr[y_i \in \Gamma(x_i)|x_i] \geq 1-\alpha \} | }{|\mathcal{S}|}\bigg] \geq 1 - \delta,
\end{align}
where we follow the convention that $0/0=0$ throughout the paper. The inner probability in (\ref{eq:batch_goal}) is taken with respect to the reference distribution of the label $y_i \sim p^*(y_i|x_i)$ given test input $x_i$ as in (\ref{eq:ideal_goal}), while the outer expectation in  (\ref{eq:batch_goal}) is evaluated with respect to the covariates $\{x_i\}_{i=1}^{|\mathcal{D}^\text{te}|}$ in the test input dataset $\mathcal{D}^\text{te}$ and over any reference data used to produce the prediction set $\Gamma(x)$ (see Sec.~\ref{subsec:edge_only} for details). The inequality (\ref{eq:batch_goal}) imposes that the fraction of edge-processed inputs for which the conditional coverage condition (\ref{eq:ideal_goal}) is met is no smaller than $1-\delta$.

Since the requirement (\ref{eq:batch_goal}) can be always guaranteed by a cascading procedure that defers all inputs to the cloud or that returns the trivial prediction set $\Gamma(x) = \mathcal{Y}$, it is important to evaluate the performance of the edge-cloud systems also in terms of the deferral rate and of the informativeness of the prediction set.

The deferral rate (DR) evaluates the expected fraction of the test samples deferred to the cloud, i.e.,
\begin{align} \label{eq:deferral_ratio}
   \text{DR} = \mathbb{E}\left[ 1 - \frac{|\mathcal{S}|}{|\mathcal{D}^{\text{te}}|} \right],
\end{align}
where the expectation is taken with respect to the distribution of the selected subset $\mathcal{S}$ and over any reference data used to generate the prediction set $\Gamma(x)$. The deferral rate (\ref{eq:deferral_ratio}) ranges from $0$, indicating that all test samples are processed at the edge, to $1$, indicating that all test inputs are deferred to the cloud.

The informativeness of the prediction set is evaluated by comparing the set size $|\Gamma(x)|$ with the cloud model's set size $|\Gamma^*(x)|$. Accordingly, the expected size of the prediction set $\Gamma(x)$ normalized by the size of the cloud prediction set $\Gamma^*(x)$, referred to normalized inefficiency (NI), is defined as
\begin{align} \label{eq:inefficiency}
     \text{NI} = \frac{1}{|\mathcal{D}^{\text{te}}|} \mathbb{E}\left[ \sum_{x_i \in \mathcal{D}^{\text{te}}} \frac{|\Gamma(x_i)|}{|\Gamma^*(x_i)|}  \right],
\end{align}
where the expectation is taken over the randomness of the covariates $\{x_i\}_{i=1}^{|\mathcal{D}^\text{te}|}$ in the test input dataset $\mathcal{D}^\text{te}$ and over any reference data used to generate the prediction set $\Gamma(x)$.

The normalized inefficiency (\ref{eq:inefficiency}) measures the relative increase in the prediction set size caused by the use of the edge model for some of the test inputs. Accordingly,  a normalized inefficiency equal to $1$ indicates an edge-cloud system that is as efficient as the cloud prediction, while a larger normalized inefficiency quantifies the loss of information about the label that is entailed by the use of the edge model.

All in all, a well-designed edge-cloud prediction mechanism should seek to minimize the deferral rate (\ref{eq:deferral_ratio}) and the normalized inefficiency (\ref{eq:inefficiency}), while satisfying the average satisfaction rate guarantee in (\ref{eq:batch_goal}).

\section{Baselines} \label{sec:baselines}

In this section, we introduce baseline prediction strategies based only on the cloud or edge models, as well as a conventional heuristic cascading strategy based on the edge model's confidence \cite{gui2024conformal}.
\vspace{-0.5cm}
\subsection{Cloud-Only Inference}
As discussed in Sec.~\ref{subsec:setting}, the cloud-only HMS $\Gamma^*(x)$ in (\ref{eq:cloud_prediction_set}) is the smallest-cardinality prediction set satisfying the conditional coverage requirement (\ref{eq:ideal_goal}). Since it satisfies (\ref{eq:ideal_goal}), it also directly meets the relaxed requirement (\ref{eq:batch_goal}) for any tolerated violation level $\delta$. Furthermore, the normalized inefficiency (\ref{eq:inefficiency}) equals $\text{NI}=1$. However, this scheme has the highest deferral rate, i.e., $\text{DR}=1$, since all inputs are escalated to the cloud.
\vspace{-0.25cm}
\subsection{Edge-Only Inference} \label{subsec:edge_only}

At the other side of the spectrum with respect to cloud-only schemes are methods that leverage only the edge model $p^e(y|x)$, without requiring access to the cloud. We review three such methods, a baseline edge-only HMS scheme, CP, and LCP. By definition, all these schemes exhibit the minimum deferral rate $\text{DR}=0$.

\subsubsection{Edge highest mass set}

When we replace cloud predictive distribution $p^*(y|x)$ with the edge predictive distribution $p^e(y|x)$ in the HMS (\ref{eq:cloud_prediction_set}), we obtain the prediction set
\begin{align} \label{eq:edge_HDS}
    \Gamma^e(x) = \arg \min_{\Gamma^e(x) \subseteq \mathcal{Y}} |\Gamma^e(x)| \text{  s.t. } p^e(\Gamma^e(x)|x) \geq 1-\alpha.
\end{align}
The performance of this prediction set is highly sensitive to the edge model's calibration performance. Over-confident edge models tend to produce excessively small edge HMS (\ref{eq:edge_HDS}), possibly with normalized inefficiency $\text{NI}< 1$, violating the target coverage constraint (\ref{eq:batch_goal}). In contrast, under-confident edge models produce excessively large, and thus very inefficient, prediction set (\ref{eq:edge_HDS}), with normalized inefficiency $\text{NI}>1$. In general, this approach does not satisfy the target coverage requirement (\ref{eq:batch_goal}). 

\subsubsection{Conformal prediction} \label{subsec:edge_CP}
To mitigate edge model miscalibration, CP leverages a held-out labeled calibration dataset $\mathcal{D}^{\text{cal}} = \{(x_i, y_i)\}_{i=1}^{|\mathcal{D}^{\text{cal}}|}$ generated from the ground-truth data distribution $p^*(x,y)$ (\ref{eq:ground-truth_dist}) to obtain a prediction set with marginal validity guarantees. 

Fix a function $V(x,y)$ measuring the discrepancy between the prediction produced by the edge model $p^e(y|x)$ and the true label $y$, such as the negative log-loss $V(x,y) = -\log p^e(y|x)$. This function is applied to all data points in the calibration dataset, producing the set of scores 
\begin{align} \label{eq:edge_CP_error_set}
    \mathcal{V} = \{ V(x_i, y_i) \}_{i=1}^{|\mathcal{D}^{\text{cal}}|}.
\end{align}

Given an input $x$, CP constructs the edge prediction set by including all the labels $y\in \mathcal{Y}$ for which the score $V(x,y)$ does not exceed a threshold $q$, i.e., 
\begin{align} \label{eq:edge_CP}
    \Gamma^e(x) = \{ y \in \mathcal{Y}: V(x,y) \leq q \}.
\end{align}
The threshold $q$ is selected as the $(1-\alpha)$-th lower quantile of the empirical distribution of the scores in set $\mathcal{V}$ (\ref{eq:edge_CP_error_set}), with a small correction, i.e.,
\begin{align} \label{eq:CP_quantile}
    q = 
    \text{Quantile}_{1-\alpha} \left( 
        \sum_{i=1}^{|\mathcal{D}^{\text{cal}}|} \frac{1}{1+|\mathcal{D}^{\text{cal}}|} \delta_{V(x_i, y_i)} 
        + \frac{1}{1+|\mathcal{D}^{\text{cal}}|} \delta_{\infty} 
    \right),
\end{align}
with $\delta_{V}$ denoting a point mass at $V$. The function $\text{Quantile}_{1-\alpha} (\cdot)$ finds the smallest value $q$ so that the total weight of samples below $q$ is at least $1-\alpha$.

CP provides only marginal validity guarantees \cite[Eq.~(1)]{angelopoulos2021gentle}, that is, the prediction set $\Gamma^e(x)$ (\ref{eq:edge_CP}) satisfies the inequality
\begin{align} \label{eq:marginal_validity}
    \Pr \left[ y \in \Gamma^e(x) \right] \geq 1-\alpha,
\end{align}
where the probability is evaluated with respect to the joint distribution $p^*(x,y)$ of the test pair $(x, y)$ and to the calibration dataset used to generate the edge prediction set $\Gamma^e(x)$. The condition (\ref{eq:marginal_validity}) is weaker than the conditional coverage requirement (\ref{eq:ideal_goal}), and thus CP does not guarantee the required inequality (\ref{eq:batch_goal}).

\subsubsection{Localized conformal prediction} \label{subsec:edge_LCP}
While CP-based methods can only guarantee the marginal coverage condition (\ref{eq:marginal_validity}), a modified version of CP, known as LCP, attempts to improve conditional coverage by selecting the threshold $q$ in (\ref{eq:CP_quantile}) as a function of the test input $x$ \cite{qkae103}.

To elaborate, fix any localization kernel, such as the Gaussian kernel
\begin{align} \label{eq:gaussian_kernel}
    H(x_1, x_2) = \exp \left( -\frac{\lVert x_1 - x_2 \rVert_2^2}{2h^2} \right),
\end{align}
with kernel bandwidth $h > 0$. Then, given a test input $x$, LCP draws a random perturbation $\tilde{x}$ of the test input $x$ by sampling from a distribution with density proportional to the kernel $H(x, \cdot)$. Then, LCP evaluates the threshold
\begin{align} \label{eq:localized_quantile}
    \hat{q}(x) = 
    \text{Quantile}_{1-\alpha} \left( 
        \sum_{i=1}^{|\mathcal{D}^{\text{cal}}|} w_{x_i}  \delta_{V(x_i, y_i)} 
        + w_{x} \delta_{\infty} 
    \right),
\end{align}
where the normalized weights are
\begin{align} \label{eq:normalized_weight}
    w_{x_i} = \frac{H(x_i, \tilde{x})}{H(x,\tilde{x}) + \sum_{i=1}^{|\mathcal{D}^{\text{cal}}|} H(x_i, \tilde{x})}, 
    \nonumber \\
    w_{x} = \frac{H(x, \tilde{x})}{H(x,\tilde{x}) + \sum_{i=1}^{|\mathcal{D}^{\text{cal}}|} H(x_i, \tilde{x})}.
\end{align}
This approach localizes the threshold (\ref{eq:localized_quantile}) around the test input $x$ by assigning higher weights to calibration points closer to $x$.

Finally, the LCP set is 
\begin{align} \label{eq:edge_LCP}
    \Gamma^e(x) = \{ y \in \mathcal{Y}: V(x,y) \leq \hat{q}(x) \}.
\end{align}
By the definition of the localized threshold $\hat{q}(x)$ (\ref{eq:localized_quantile}), a small kernel bandwidth $h$ yields more localized prediction sets, while a large kernel bandwidth $h$ reduces LCP to CP (\ref{eq:edge_CP}).

Although there is numerical evidence that LCP can enhance conditional coverage over CP \cite[Thm. 2]{qkae103}, it still guarantees only the marginal validity condition (\ref{eq:marginal_validity}) \cite[Thm. 1]{qkae103}, not meeting the target requirement (\ref{eq:batch_goal}).

\subsection{Confidence-Based Model Cascading} \label{sec:CbD}

As seen, edge-only schemes can not offer the target conditional coverage guarantees (\ref{eq:batch_goal}). In this subsection, we review conventional edge-cloud systems in which the deferral option is implemented by following a heuristic confidence-based rule \cite{gui2024conformal}. 

Given an input $x$, the edge system evaluates a measure of confidence on its output, and decides to defer the decision to the cloud when the confidence level is below a pre-determined threshold. In this work, we adopt the common top-$1$ confidence measure, i.e., $\max_{y \in \mathcal{Y}} p^e(y|x)$ \cite{gui2024conformal}. Accordingly, the edge-cloud system produces the prediction sets based on the rule
\begin{align} \label{eq:conventional_cascade}
       \Gamma(x) =
        \begin{cases}
       \Gamma^*(x), & \text{if $ \max_{y \in \mathcal{Y}} p^e(y|x) < \gamma$ }  \\
       \Gamma^e(x), & \text{if $\max_{y \in \mathcal{Y}} p^e(y|x) \geq \gamma$ },
        \end{cases}
\end{align}
with a pre-determined threshold $\gamma \in [0,1]$, where $\Gamma^e(x)$ is an edge-only prediction set, such as HMS (\ref{eq:edge_HDS}), CP (\ref{eq:edge_CP}), or LCP (\ref{eq:edge_LCP}).

The threshold $\gamma$ is typically selected as $\gamma=1-\delta$ \cite{gui2024conformal}. This way, the edge-only prediction sets for which the edge confidence exceeds the target average satisfaction level $1-\delta$ in (\ref{eq:batch_goal}) are processed by the edge model, while others are outsourced to the cloud.

\textbf{Remark 2:}  The confidence-based deferral (CbD) rule (\ref{eq:conventional_cascade}) escalates to the cloud on the basis of the maximum edge confidence level, a typical approach in uncertainty-aware routing \cite{jitkrittum2023does, fithian2014optimal, rabanser2025gatekeeper}. Moreover, CP (\ref{eq:edge_CP}) and LCP (\ref{eq:edge_LCP}) calibrate the edge prediction set $\Gamma^e(x)$ based on the calibration dataset $\mathcal{D}^{\text{cal}}$, making CbD+CP and CbD+LCP instances of calibration-aware cascaded inference \cite{schuster2022confident, jung2024trust}. None of these schemes provides the formal conditional reliability guarantee required by \eqref{eq:batch_goal}. \hfill $\blacksquare$

\section{Conformal Alignment-based Cascading} \label{sec:CA}

In this section, we introduce a CAb model cascading mechanism that provably meets the target coverage requirement (\ref{eq:batch_goal}). To this end, we formulate the escalation procedure as a MHT problem by tailoring the CA method \cite{gui2024conformal} to adopt the conditional coverage probability (\ref{eq:ideal_goal}) as the alignment score.

\subsection{Model Cascading via Multiple Hypothesis Testing}
The proposed CAb methodology is based on the observation that the requirement (\ref{eq:batch_goal}) can be interpreted as a FDR constraint in an MHT procedure \cite{sedgwick2014understanding}. To elaborate, given any edge-only prediction set $\Gamma^e(x)$, such as HMS (\ref{eq:edge_HDS}), CP (\ref{eq:edge_CP}), or LCP (\ref{eq:edge_LCP}), we write the conditional coverage probability as
\begin{align} \label{eq:alignment_score}
      C^*(x) = p^*(\Gamma^e(x)|x).
\end{align}
In the following, we interpret the probability $C^*(x)$ as an alignment score, measuring how well the edge-only prediction set $\Gamma^e(x)$ aligns with the oracle prediction set in $\Gamma^*(x)$ (\ref{eq:cloud_prediction_set}). In particular, if the edge-only prediction set $\Gamma^e(x)$ aligns well with the cloud-only HMS $\Gamma^*(x)$, the alignment score must be no smaller than the target conditional coverage probability $1-\alpha$.

For any test input $x_i \in \mathcal{D}^{\text{te}}$, we wish to decide whether the edge model prediction set meets the conditional coverage requirement (\ref{eq:ideal_goal}). To formalize this problem, we assign each test input $x_i \in \mathcal{D}^{\text{te}}$ to a null hypothesis $\mathcal{H}_i$ that the edge-only prediction set $\Gamma^e(x)$ fails to satisfy the conditional coverage requirement (\ref{eq:ideal_goal}). This can be expressed mathematically via the inequality
\begin{align} \label{eq:null_hypothesis}
    \mathcal{H}_i: C^*(x_i) < 1-\alpha.
\end{align} 
While the hypothesis (\ref{eq:null_hypothesis}) pertains to an individual test input $x_i \in \mathcal{D}^{\text{te}}$, the average satisfaction rate guarantee (\ref{eq:batch_goal}) requires the simultaneous consideration of the hypothesis for all test inputs $x_i \in \mathcal{D}^{\text{te}}$, inducing an MHT problem.
\begin{figure*} [tb] 
    \centering
    \centerline{\includegraphics[scale=0.22]{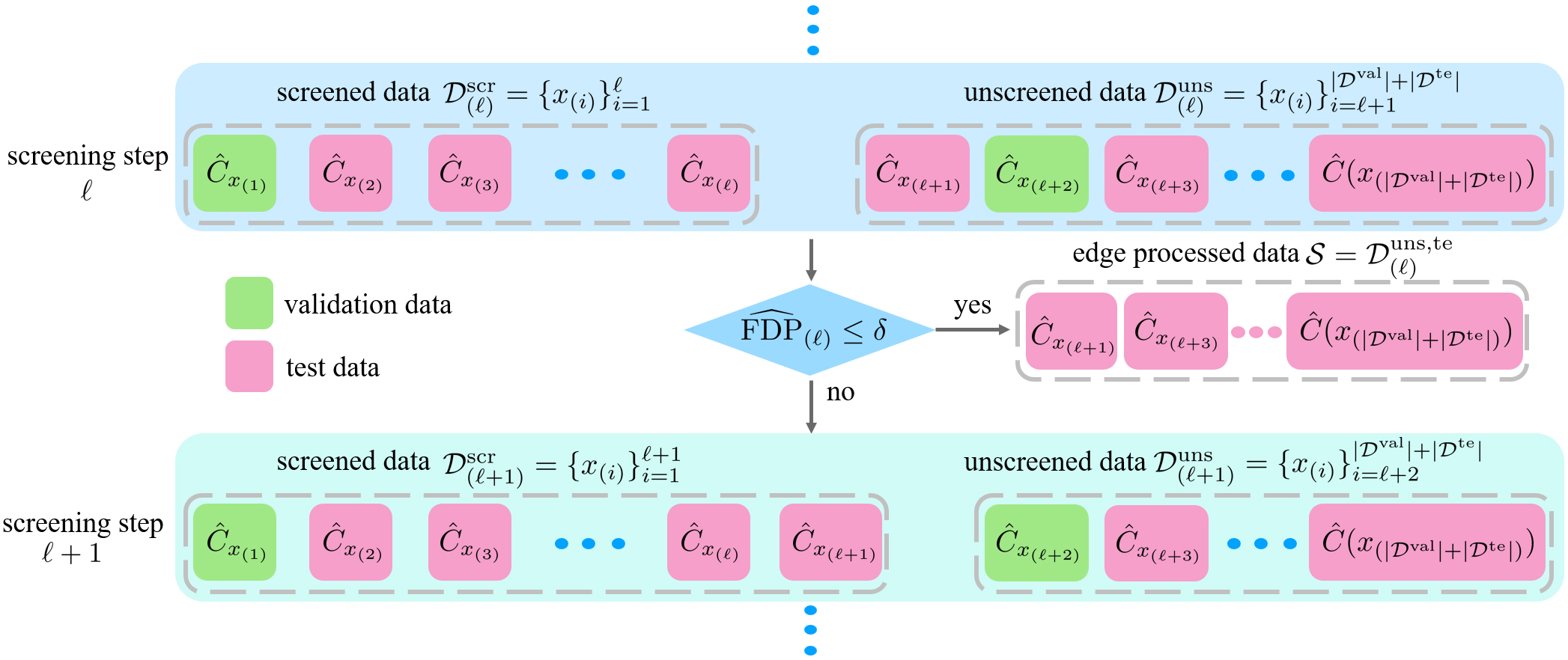}}
     \caption{The proposed CAb model cascading adopts a sequential screening methodology to ensure that the edge-processed subset $\mathcal{S}$ satisfies the constraint (\ref{eq:batch_goal}). This procedure evaluates inputs in the joint test and validation dataset $\mathcal{D}^{\text{te}} \cup \mathcal{D}^{\text{val}}$ in the order (\ref{eq:screen_order}) of increasing estimate alignment score $\hat{C}(x)$. At each step, the CAb method estimates the false discovery proportion (FDP) of the unscreened test inputs based on the unscreened validation data using (\ref{eq:FDP_estimator}). The procedure terminates at the earliest step $\ell_{\text{CA}}$ in (\ref{eq:stopping_rule}) when the estimated FDP of unscreened test inputs falls below the pre-determined tolerated level $\delta$.}
    \label{fig:ACS} 
\end{figure*}
In this MHT problem, the subset $\mathcal{S} \subseteq \mathcal{D}^{\text{te}}$ of test inputs that are processed at the edge corresponds to the subset of null hypotheses $\{ \mathcal{H}_i\}_{i=1}^{|\mathcal{D}^{\text{te}}|}$ in (\ref{eq:null_hypothesis}) that are rejected. Accordingly, we can reformulate the edge-cloud prediction set $\Gamma(x_i)$ in (\ref{eq:original_prediction}) as
\begin{align} \label{eq:final_prediction}
       \Gamma(x_i) =
        \begin{cases}
       \Gamma^*(x_i), & \text{if input $x_i \notin \mathcal{S}$, i.e., $\mathcal{H}_i$ is accepted}  \\
       \Gamma^e(x_i) , & \text{if input $x_i \in \mathcal{S}$, i.e., $\mathcal{H}_i$ is rejected}.
        \end{cases}
\end{align}

Furthermore, the average satisfaction rate guarantee (\ref{eq:batch_goal}) can be expressed in terms of the false discovery proportion (FDP), i.e., the fraction of test samples in the edge-processed subset $\mathcal{S}$ for which the null hypothesis (\ref{eq:null_hypothesis}) is incorrectly rejected \cite{jin2023selection}. By the definition of the null hypothesis in (\ref{eq:null_hypothesis}), the FDP is defined as 
\begin{align} \label{eq:FDP}
   \text{FDP}(\mathcal{S}) = 
\frac{\left| \{ x_i \in \mathcal{S} : C^*(x_i) < 1 - \alpha \} \right|}
{|\mathcal{S}|}.
\end{align}
Then, the average satisfaction rate guarantee (\ref{eq:batch_goal}) can be equivalently written as the inequality
\begin{align} \label{eq:batch_goal_reformulated}
    \text{FDR} = \mathbb{E}[\text{FDP}(\mathcal{S})] \leq \delta,
\end{align}
where the expectation is evaluated with respect to both the distribution of the subset $\mathcal{S}$ and the reference data used to generate the prediction set $\Gamma(x)$. The expected value of the FDP in (\ref{eq:batch_goal_reformulated}) is known as the FDR. 

\subsection{Conformal Alignment-Based Model Cascading}

In this subsection, we describe the proposed CAb model cascading procedure that enforces the constraint (\ref{eq:batch_goal_reformulated}). As illustrated in Fig.~\ref{fig:ACS}, we adopt a sequential screening approach \cite{mary2022semi, gui2025acs}, whereby the inputs that are likely to violate the coverage condition (\ref{eq:ideal_goal}) are progressively eliminated until the remaining unscreened test inputs satisfy the requirement (\ref{eq:batch_goal_reformulated}). 

The CAb method assumes the availability of a reference dataset $\mathcal{D}$ consisting of pairs $(x, C^*(x))$, where $C^*(x)$ is the true alignment score (\ref{eq:alignment_score}). Note that the label $C^*(x)$ is obtained by querying the cloud model during an offline phase. The reference dataset $\mathcal{D}$ is partitioned into two disjoint datasets, i.e., a training dataset $\mathcal{D}^{\text{tr}}$ and a validation dataset $\mathcal{D}^{\text{val}}$, which are used as detailed next.

Since the true alignment score $C^*(x)$ is not available for test inputs, we introduce an alignment score predictor $\hat{C}(x)$ \cite{jin2023selection, gui2024conformal}. This predictor is trained on the training dataset $\mathcal{D}^{\text{tr}}= \{ (x_i, C^*(x_i)) \}_{i=1}^{|\mathcal{D}^{\text{tr}}|}$ in an offline phase using any supervised learning method. No specific assumption is imposed on the quality of this predictor.

\subsubsection{Sequential Screening}
Given a pre-trained alignment score predictor $\hat{C}(x)$, and given an input batch $\mathcal{D}^{\text{te}}$, the proposed CAb methodology uses the validation dataset $\mathcal{D}^{\text{val}} = \{ (x_i, C^*(x_i)) \}_{i=1}^{|\mathcal{D}^{\text{val}}|}$ to guide the selection of the subset $\mathcal{S} \subseteq \mathcal{D}^{\text{te}}$ of test inputs to process at the edge. To start, the test dataset $\mathcal{D}^{\text{te}}$ is combined with the validation dataset $\mathcal{D}^{\text{val}}$ to form a joint set
\begin{align}
    \mathcal{D}^{\text{te}} \cup \mathcal{D}^{\text{val}} = \{x_1, \cdots, x_{|\mathcal{D}^{\text{val}}|}, \cdots, x_{|\mathcal{D}^{\text{val}}|+|\mathcal{D}^{\text{te}}|} \},
\end{align} 
where $\{x_1, \cdots, x_{|\mathcal{D}^{\text{val}}|} \}$ represents the inputs for the labeled validation samples in the validation dataset $\mathcal{D}^{\text{val}}$. Then, the edge device ranks all samples in the union $\mathcal{D}^{\text{te}} \cup \mathcal{D}^{\text{val}}$ as $x_{(1)}, \cdots, x_{(|\mathcal{D}^{\text{val}}|+|\mathcal{D}^{\text{te}}|)}$ in ascending order of their predicted alignment score $\hat{C}(x)$, i.e.,
\begin{align} \label{eq:screen_order}
    \hat{C}(x_{(1)}) \leq \cdots \leq \hat{C}(x_{(|\mathcal{D}^{\text{val}}|)}) \leq \cdots \leq \hat{C}(x_{(|\mathcal{D}^{\text{val}}|+|\mathcal{D}^{\text{te}}|)}).
\end{align}
Intuitively, this step lists the inputs in order from least to most promising for edge processing.

As illustrated in Fig.~\ref{fig:ACS}, the edge proceeds to screen the data points in the joint dataset $\mathcal{D}^{\text{te}} \cup \mathcal{D}^{\text{val}}$ following the order in (\ref{eq:screen_order}), with screening steps indexed by an integer $\ell =1,2,\cdots, |\mathcal{D}^{\text{val}}|+|\mathcal{D}^{\text{te}}|$. Accordingly, at each screening step $\ell$, we screen the new input $x_{(\ell)}$, and we define the screened inputs and the unscreened inputs as 
\begin{align} \label{eq:unscreend_test_inputs}
     \mathcal{D}^{\text{scr}}_{(\ell)} = \{x_{(i)}\}_{i=1}^{\ell} \quad \text{and}  \quad \mathcal{D}^{\text{uns}}_{(\ell)} = \{x_{(i)}\}_{i=\ell + 1}^{|\mathcal{D}^{\text{val}}|+|\mathcal{D}^{\text{te}}|},
\end{align}
respectively. Since the unscreened input subset $\mathcal{D}^{\text{uns}}_{(\ell)}$ generally includes both validation and test data, we also partition this set into unscreened validation and test subsets as
\begin{align} \label{eq:selected_set}
    \mathcal{D}^{\text{uns,val}}_{(\ell)}  =\mathcal{D}^{\text{uns}}_{(\ell)} \cap \mathcal{D}^{\text{val}} \quad \text{and}\quad
    \mathcal{D}^{\text{uns,te}}_{(\ell)}  =\mathcal{D}^{\text{uns}}_{(\ell)} \cap \mathcal{D}^{\text{te}},
\end{align}
respectively. Initially, at step $\ell = 0$, the screened input subset is an empty set, $\mathcal{D}^{\text{scr}}_{(0)} = \emptyset$.

The screening procedure proceeds along steps $\ell =1,2,\cdots, |\mathcal{D}^{\text{val}}|+|\mathcal{D}^{\text{te}}|$, until a certain condition is met at some step $\ell_{\text{CA}} \leq |\mathcal{D}^{\text{val}}|+|\mathcal{D}^{\text{te}}|$. Once this occurs, the CAb procedure returns the set 
\begin{align} \label{eq:edge_process_set}
    \mathcal{S} = \mathcal{D}^{\text{uns,te}}_{(\ell_{\text{CA}})}
\end{align}
of unscreened test inputs. By the ordering (\ref{eq:screen_order}), this set contains all test inputs $x_i \in \mathcal{D}^{\text{te}}$ with an estimated alignment score $\hat{C}(x)$ no smaller than $\hat{C}(x_{(\ell_{\text{CA}})})$, i.e.,
\begin{align}
    \mathcal{S} = \{ x_i \in  \mathcal{D}^{\text{te}}: \hat{C}(x_i) \geq \hat{C}(x_{(\ell_{\text{CA}})})\}.
\end{align}

To determine the stopping time $\ell_{\text{CA}}$, as illustrated in Fig.~\ref{fig:ACS}, at each step $\ell$, the CAb method estimates the FDP (\ref{eq:FDP}) of the subset of unscreened test inputs $\mathcal{D}^{\text{uns,te}}_{(\ell)}$ by using the corresponding FDP of the subset of unscreened validation inputs $\mathcal{D}^{\text{uns,val}}_{(\ell)}$ as
\begin{align} \label{eq:FDP_estimator}
    \widehat{\text{FDP}}_{( \ell)} = \frac{|\mathcal{D}^{\text{te}}|}{1 + |\mathcal{D}^{\text{val}}|} \frac{1+| \{x_i \in \mathcal{D}^{\text{uns,val}}_{( \ell)}: C^*(x_i) < 1-\alpha      \}|}{|\mathcal{D}^{\text{uns,te}}_{(\ell)}|}.
\end{align}
Intuitively, the multiplicative term $|\mathcal{D}^{\text{te}}| / (1 + |\mathcal{D}^{\text{val}}|)$ in (\ref{eq:FDP_estimator}) compensates for the discrepancy in the sizes of the validation and the test dataset \cite[Eq.~(2)]{gui2025acs}. Note that the FDP for the unscreened validation inputs, which is obtained as the ratio $| \{ x_i \in \mathcal{D}^{\text{uns,val}}_{(\ell)}: C^*(x_i) < 1 - \alpha \} | / |\mathcal{D}^{\text{uns,val}}_{(\ell)}|$ by the definition (\ref{eq:FDP}), can be evaluated since the ground-truth alignment scores $C^*(x_i)$ are available for the validation samples $x_i \in \mathcal{D}^{\text{val}}$. 

With this estimate, the CAb method terminates the sequential screening procedure at the first step that meets the condition (\ref{eq:batch_goal_reformulated}), with the estimate (\ref{eq:FDP_estimator}) used in lieu of the true FDP, i.e.,
\begin{align} \label{eq:stopping_rule}
    \ell_{\text{CA}} = \inf \{ \ell \geq 0: \widehat{\text{FDP}}_{( \ell)} \leq \delta \}.
\end{align}
As discussed next, this procedure satisfies the requirement (\ref{eq:batch_goal}).

\subsubsection{Theoretical Guarantees}
The output $\mathcal{S} = \mathcal{D}_{(l_\text{CA})}^\text{uns,te}$ of the CAb model cascading methodology satisfies the target FDR constraint (\ref{eq:batch_goal_reformulated}), which coincides with the target average satisfaction rate guarantee (\ref{eq:batch_goal}).

\textbf{Proposition 1:} \textit{If the examples in the reference dataset $\mathcal{D}$, and the test dataset $\mathcal{D}^{\text{te}}$ are exchangeable, then, for any pre-determined average satisfaction level $1-\delta \in [0,1]$, the output subset $\mathcal{S}$ in (\ref{eq:edge_process_set}) satisfies the average satisfaction rate requirement (\ref{eq:batch_goal}).
}

\emph{Proof}: The proof of this proposition follows directly from the FDR control of CA (see \cite[Thm.~1]{gui2025acs} for details). For completeness, a proof tailored to the sequential screening procedure presented in this subsection, which was introduced in \cite{gui2025acs}, can be found in the Appendix. \hfill $\blacksquare$
\vspace{-0.2cm}
\section{Experiments} \label{sec:results}

In this section, to validate the proposed approaches, we report empirical results for vision and QA tasks.
\subsection{Performance Metrics}
For both tasks, we consider the following evaluation metrics:
\begin{itemize}
    \item {Average satisfaction rate}, the average proportion of edge-processed test samples whose conditional coverage probability (\ref{eq:idea_goal_reformula}) is no smaller than the desired requirement $1-\alpha$, estimating the left-hand side of (\ref{eq:batch_goal}).
    \item {Deferral rate}, the averaged fraction of test samples deferred to the cloud, estimating (\ref{eq:deferral_ratio}).
    \item {Normalized inefficiency}, the expected size of the prediction set $\Gamma(x)$ normalized by the size of the oracle cloud prediction set $\Gamma^*(x)$, estimating (\ref{eq:inefficiency}).
\end{itemize}
\subsection{Implementation}
The calibration dataset $\mathcal{D}^{\text{cal}} = \{(x_i, y_i)\}_{i=1}^{|\mathcal{D}^{\text{cal}}|}$ is used by the edge model $p^e(y|x)$ to construct the edge prediction set $\Gamma^e(x)$ as in (\ref{eq:edge_CP}) or in (\ref{eq:edge_LCP}). We randomly partition the reference dataset $\mathcal{D}$ into two disjoint datasets, namely the training dataset $\mathcal{D}^{\text{tr}} = \{ (x_i, C^*(x_i)) \}_{i=1}^{|\mathcal{D}^{\text{tr}}|}$ used to train the alignment score predictor $\hat{C}(x)$ in (\ref{eq:screen_order}), and the validation dataset $\mathcal{D}^{\text{val}} = \{ (x_i, C^*(x_i)) \}_{i=1}^{|\mathcal{D}^{\text{val}}|}$ used for the CAb deferral decision in (\ref{eq:stopping_rule}). We fix the sizes for each dataset as $|\mathcal{D}^{\text{cal}}| = 500$, $|\mathcal{D}^{\text{tr}}|=200$, $|\mathcal{D}^{\text{val}}|=500$, and $|\mathcal{D}^{\text{te}}|=100$. For the alignment score predictor $\hat{C}(x)$ in (\ref{eq:screen_order}), we adopt a regression model that takes as input the probability $p^e(\Gamma^e(x)|x)$, which represents the conditional coverage probability for the edge prediction set $\Gamma^e(x)$ as estimated by the edge model.

All results are averaged over $200$ independent runs, with each run corresponding to an independent split of the datasets. All the experiments are implemented via PyTorch \cite{paszke2019pytorch}, and run over a GPU server with a single NVIDIA A100 card\footnote{Code can be found at \url{https://github.com/kclip/Edge-Cloud-Conformal-Alignment}.}.

\subsection{Image Classification}

For the image classification task, we use the CIFAR-100 dataset \cite{krizhevsky2010cifar}, with a Bayesian WideResNet-40-2 network \cite{huang2024calibrating} and a standard WideResNet-40-2 model \cite{zagoruyko2016wide} as the cloud and edge models, respectively. The alignment score predictor $\hat{C}(x)$ is trained via XGBoost \cite{gui2024conformal}.

\begin{figure*} [htb] 
    \centering
    \centerline{\includegraphics[scale=0.25]{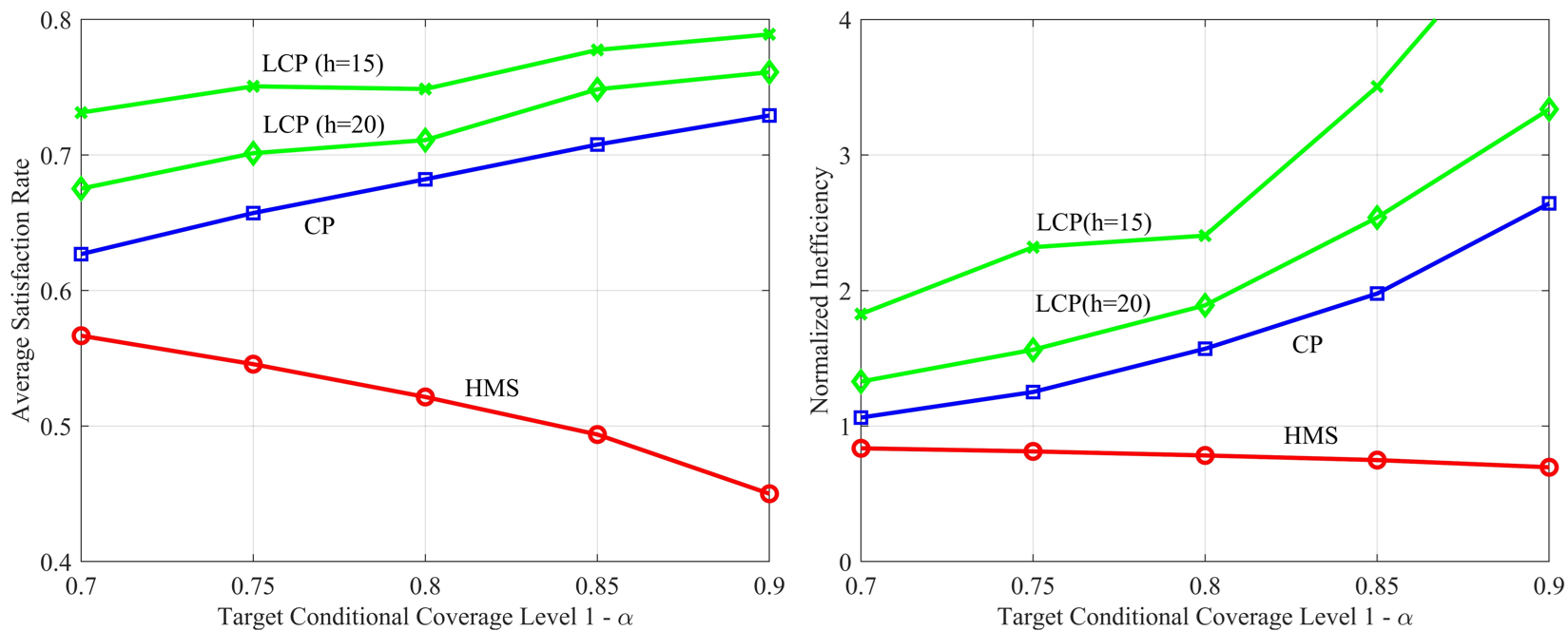}}
     \caption{Average satisfaction rate (left) and normalized inefficiency (right) versus target conditional coverage levels $1-\alpha \in \{0.7, 0.75, 0.8, 0.85, 0.9\}$ on the CIFAR-100 dataset, considering edge-only schemes: highest mass set (HMS) in (\ref{eq:edge_HDS}), conformal prediction (CP) in (\ref{eq:edge_CP}), and localized conformal prediction (LCP) in (\ref{eq:edge_LCP}) with Gaussian kernel bandwidth $h=15$ and $h=20$, respectively.}
     \vspace{-0.5cm}
    \label{fig:image_CD} 
\end{figure*}

\noindent \textbf{Empirical average satisfaction rates for edge-only schemes.} To start, we present reference results for edge-only schemes in Fig.~\ref{fig:image_CD}, where we report the average satisfaction rate and the normalized inefficiency at target conditional coverage levels $1-\alpha \in \{0.7, 0.75,0.8,0.85,0.9 \}$ for HMS (\ref{eq:edge_HDS}), CP (\ref{eq:edge_CP}), and LCP (\ref{eq:edge_LCP}). For LCP, we fix the Gaussian kernel bandwidth to $h=15$ and to $h=20$, respectively. It is emphasized that the performance of edge-only schemes in terms of conditional coverage can only be evaluated using empirical means, as edge-only schemes do not offer any formal mechanism to control the average satisfaction rate as in (\ref{eq:batch_goal}). That said, the results in Fig.~\ref{fig:image_CD} provide useful benchmarks for the cloud-aided cascading techniques studied in this work.

The empirical results in Fig.~\ref{fig:image_CD} show that edge-only schemes achieve low values of the average satisfaction rate, e.g., LCP with bandwidth $h=15$ obtaining average satisfaction rate $0.71$ and $0.79$ at target level $1-\alpha=0.7$ and $1-\alpha=0.9$, respectively. LCP with bandwidth $h=15$ attains a higher satisfaction rate than any other edge-only scheme, including LCP with bandwidth $h=20$, but at the cost of much larger prediction sets. This indicates that a more localized kernel (\ref{eq:gaussian_kernel}) helps enhance the conditional coverage by increasing the size of the prediction sets. Based on this observation, in the following, we set the bandwidth of LCP to $h=20$ to balance conditional coverage and inefficiency.

\begin{figure} [htb] 
    \centering
    \vspace{-0.25cm}
    \centerline{\includegraphics[scale=0.23]{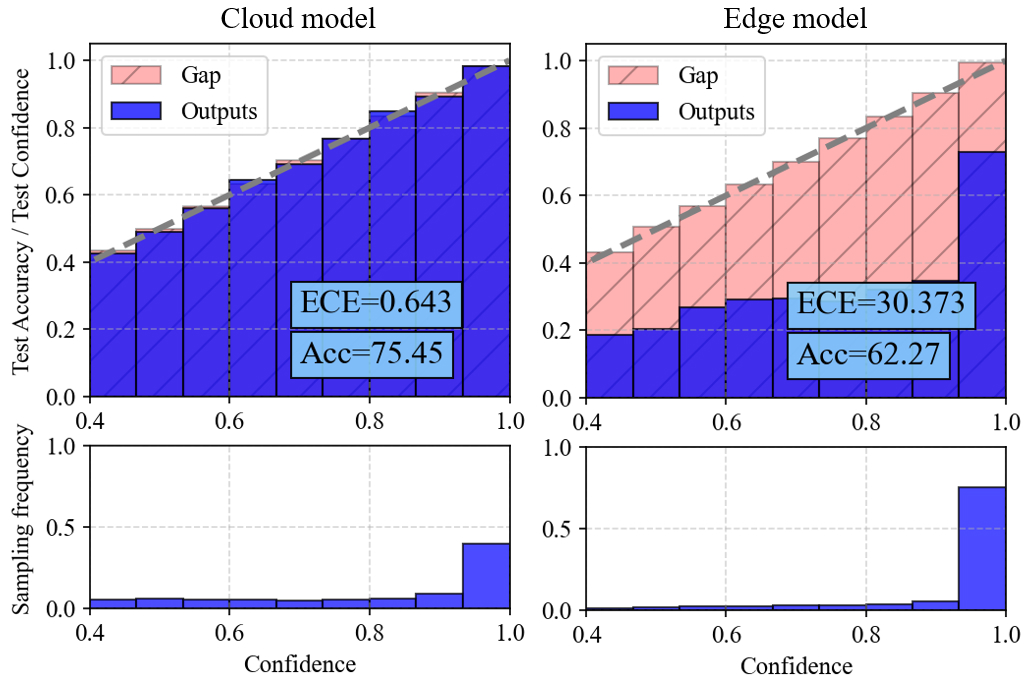}}
    \caption{Reliability diagram for the Bayesian WideResNet-40-2 cloud model (left) and the WideResNet-40-2 edge model (right), evaluated on the CIFAR-100 dataset.}
    \label{fig:image_rd} 
\end{figure}

To understand why the edge model tends to undercover the true conditional distribution of the output label in this setting, Fig.~\ref{fig:image_rd} shows the reliability diagram \cite{guo2017calibration} of the cloud and the edge model, respectively. Reliability diagrams plot average test accuracy for test samples to which the model assigns confidence levels within the same bin. Miscalibration can be measured by the size of the gaps between the dashed line, corresponding to confidence levels matching true accuracy, and the height of the test accuracy bars. The diagram in Fig.~\ref{fig:image_rd} demonstrates that the WideResNet-40-2 edge model is highly over-confident, having large positive gaps between accuracy and confidence, while the Bayesian WideResNet-40-2 cloud model is well-calibrated, with negligible discrepancy between accuracy and confidence. This implies that the edge predictive distribution $p^e(y|x)$ tends to be highly peaked around the top-$1$ label, leading to an excessively small HMS (\ref{eq:edge_HDS}). Consequently, the average satisfaction rate for HMS is seen to decline as the target coverage requirement $1-\alpha$ increases, i.e., becomes stricter, in Fig~\ref{fig:image_CD}.


While CP and LCP guarantee only marginal validity \cite{qkae103}, Fig.~\ref{fig:image_CD} demonstrates that they generally improve the achievable satisfaction rate for conditional coverage as compared to HMS. This is done by suitably increasing the size of the prediction set (see right panel of Fig.~\ref{fig:image_CD}). In particular, LCP achieves a satisfaction rate higher than CP at the expense of further inflating the prediction sets. However, as mentioned, even with LCP with a small bandwidth $h=15$, the average satisfaction rate remains quite low as a result of the poor calibration of the edge model.

\begin{figure*} [!tb] 
    \centering
    \centerline{\includegraphics[width=\textwidth]{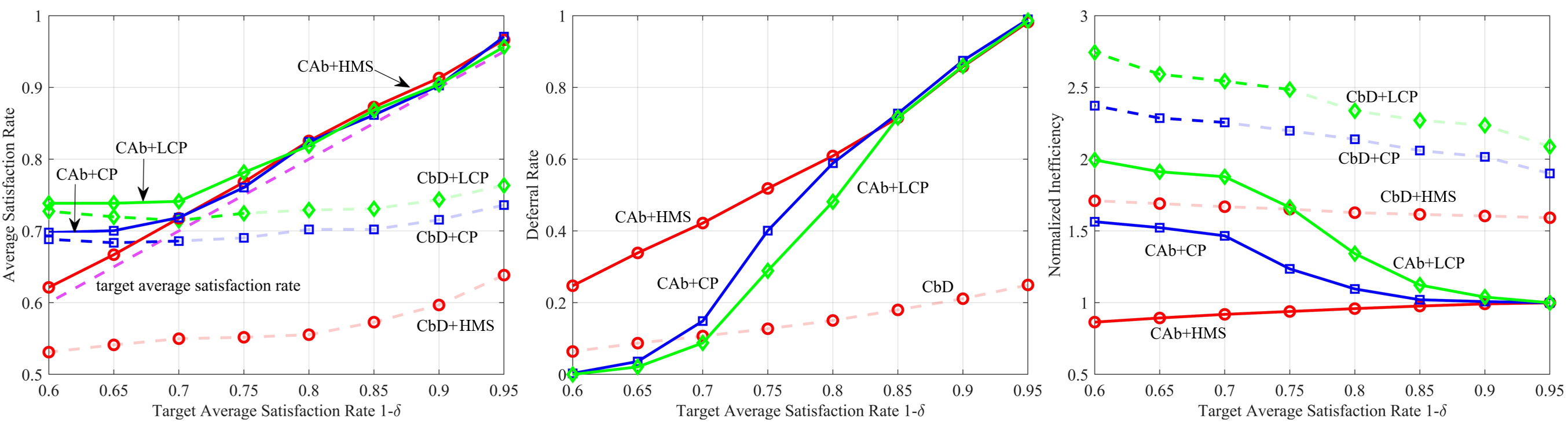}}
    \caption{Average satisfaction rate (left), deferral rate (middle), and normalized inefficiency (right) for conventional confidence-based deferral (CbD) schemes and the proposed CAb schemes versus different target average satisfaction rates $1-\delta \in \{ 0.6, 0.65, \dots, 0.95\}$ on the CIFAR-100 dataset for required conditional coverage level $1-\alpha=0.8$. The dashed straight line in the left figure indicates the target average satisfaction rate $1-\delta$. The other dashed lines in the figure represent the CbD schemes, with transparent segments indicating regimes in which CbD does not meet the target average satisfaction requirements (\ref{eq:batch_goal}).}
    \vspace{-0.5cm}
    \label{fig:image_CA} 
\end{figure*}

\noindent \textbf{Confidence-based versus conformal alignment-based model cascading.} As discussed, edge-only schemes, such as HMS, CP, and LCP, do not offer any formal mechanism to enforce a constraint on the average satisfaction rate as in (\ref{eq:batch_goal}). This requirement can only be met by implementing a deferral option to outsource inference to the cloud. To elaborate on the relative merits of different cascading techniques, we now compare the performance of CbD schemes, which operate according to the heuristic rule (\ref{eq:conventional_cascade}), to the proposed CAb schemes, which operate as detailed in Sec.~\ref{sec:CA}. We emphasize that only CAb schemes can formally guarantee the average satisfaction rate constraint (\ref{eq:batch_goal}). In this analysis, we vary the target average satisfaction level in the set $1-\delta \in \{ 0.6, 0.65, \dots, 0.95\}$, with a fixed target conditional coverage requirement $1-\alpha = 0.8$. Following a conventional thresholding strategy, we set the confidence threshold for the deferral rule (\ref{eq:conventional_cascade}) of CbD as $\gamma = 1-\delta$ (see Sec.~\ref{sec:CbD}).

Fig.~\ref{fig:image_CA} reports the average satisfaction rate, deferral rate, and the normalized inefficiency as a function of the required average satisfaction rate $1-\delta$. CbD schemes, due to the heuristic nature of the deferral rule (\ref{eq:conventional_cascade}), do not generally meet the average satisfaction requirement (\ref{eq:batch_goal}). In contrast, as formalized by Proposition 1, CAb schemes can always guarantee the condition (\ref{eq:batch_goal}), regardless of the choice of the edge prediction set, namely HMS, CP, or LCP.

For a fixed target average satisfaction rate $1-\delta$, the choice of the edge prediction set strategy, namely HMS, CP, or LCP, determines different trade-offs between deferral rates and normalized inefficiency. In particular, as seen in Fig.~\ref{fig:image_CD}, HMS yields smaller prediction sets than CP, which in turn produces smaller prediction sets than LCP. This ensures that the deferral rate decreases when switching from HMS to CP and from CP to LCP. Overall, HMS yields the smallest prediction sets with the largest deferral rate, while LCP produces the largest prediction sets with the smallest deferral rate. CP offers an intermediate solution in terms of the trade-off between deferral rate and prediction set size.

\begin{figure} [!hb] 
    \centering
    \vspace{-0.25cm}
    \centerline{\includegraphics[scale=0.21]{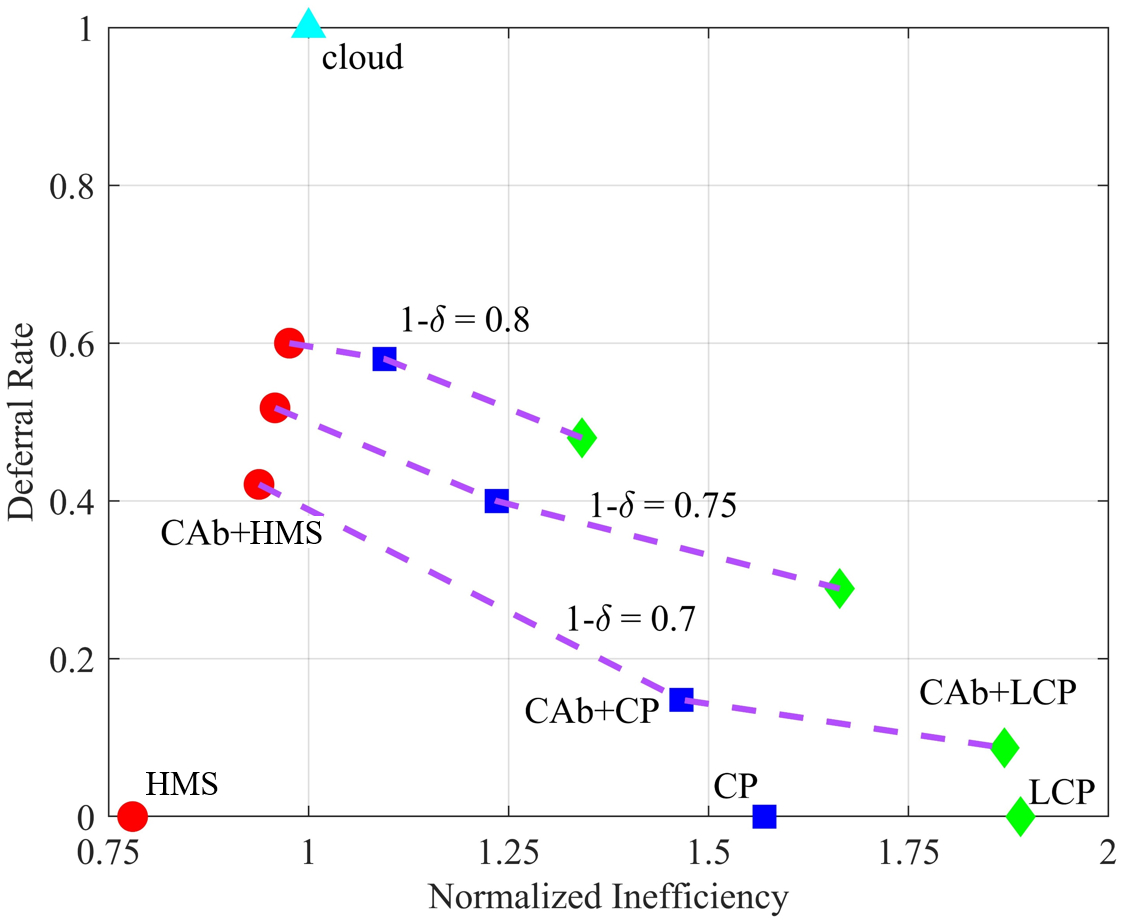}}
    \caption{Deferral rate versus normalized inefficiency obtained by changing the target average satisfaction level $1-\delta$ on CIFAR-100 dataset for CAb schemes, namely CAb+HMS, CAb+CP, and CAb+LCP. Points on the same dashed line share the same target average satisfaction rate $1-\delta$.}
    \label{fig:image_pareto} 
    \vspace{-0.3cm}
\end{figure}
\noindent \textbf{Trade-off between deferral rate and normalized inefficiency.} The trade-offs between the deferral rate and prediction set size observed in Fig.~\ref{fig:image_CA} are further analyzed in Fig.~\ref{fig:image_pareto}, which plots the deferral rate versus the normalized inefficiency for different target average satisfaction rate $1-\delta$ in (\ref{eq:batch_goal}). We focus on CAb schemes given their capacity to guarantee the average satisfaction rate constraint (\ref{eq:batch_goal}). First, we note that increasing the value of the requirement $1-\delta$ consistently raises the deferral rate for all schemes, while driving normalized inefficiency toward $1$. Furthermore, for a fixed value of the requirement $1-\delta$, larger prediction sets achieve lower deferral rates, with HMS, CP, and LCP yielding increasingly large prediction sets.

\subsection{Radio Signal Classification}
\begin{figure} [tb] 
    \centering
    \centerline{\includegraphics[scale=0.23]{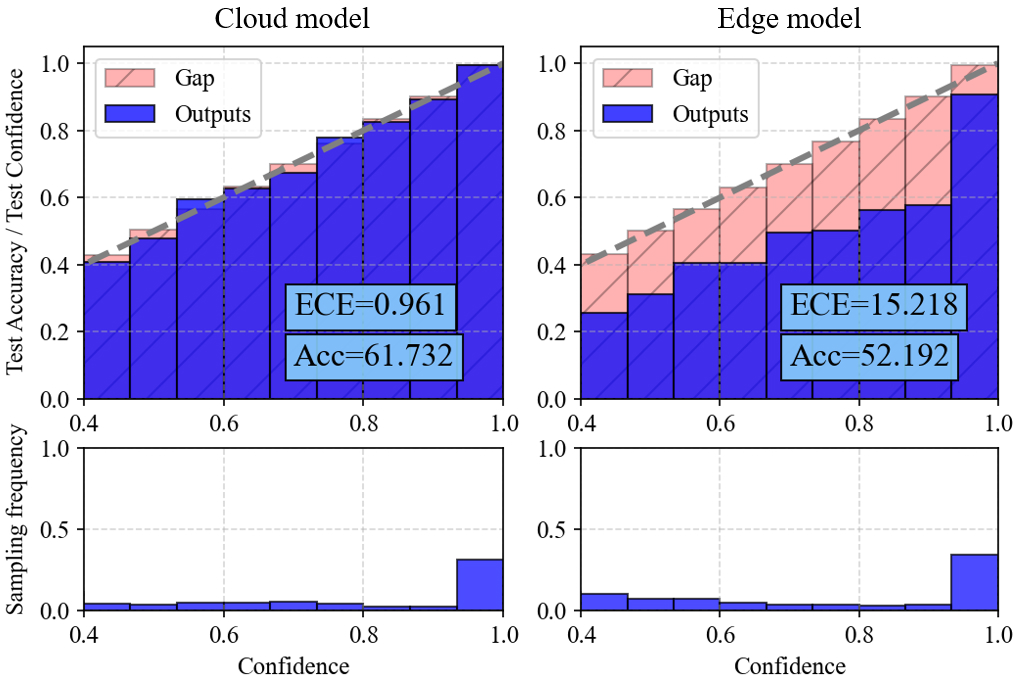}}
    \caption{Reliability diagram for the TLDNN cloud model (left) and the LSTM-DAE edge model (right), evaluated on the RadioML2018.01a dataset\cite{o2018over}.}
    \vspace{-0.3cm}
    \label{fig:radio_rd} 
\end{figure}

\begin{figure*} [!b] 
    \centering
    \centerline{\includegraphics[width=\textwidth]{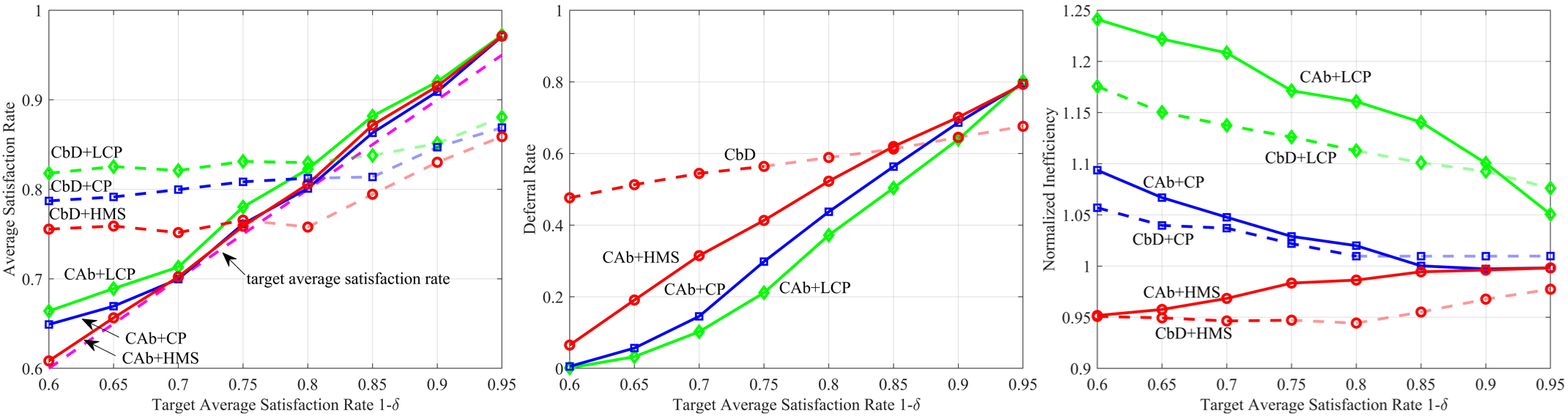}}
    \caption{Average satisfaction rate (left), deferral rate (middle), and normalized inefficiency (right) for CbD schemes and CAb schemes versus different target average satisfaction levels $1-\delta \in \{ 0.6, 0.65, \dots, 0.95\}$ on the RadioML2018.01a dataset for required conditional coverage level $1-\alpha=0.8$. The dashed straight line in the figure indicates the target average satisfaction rate $1-\delta$. The other dashed lines in the figure represent the CbD schemes. }
    \label{fig:radio_CA} 
\end{figure*}

We now turn to the radio signal classification task, which requires the identification of the modulation scheme from a received radio baseband signal. We adopt the RadioML2018.01a dataset \cite{o2018over,wang2022msmcnet}, a dataset comprising $2,555,904$ samples spanning different signal-to-noise ratio (SNR) levels from $-20$ dB to $30$ dB, with each sample corresponding to a baseband signal of length $1024$ complex numbers and a label representing one out of $24$ possible analog or digital modulation types. We focus on the low-SNR samples at $0$ dB, targeting a practical and challenging setting.

The edge and cloud models are instantiated as the LSTM-DAE network \cite{ke2021real} and the TLDNN network \cite{qu2024enhancing}, respectively. LSTM-DAE is a lightweight recurrent architecture with approximately $15$K parameters, whereas the TLDNN is a transformer-based network with $276.66$K parameters, achieving stronger recognition and reliability performance. The alignment score predictor $\hat{C}(x)$ is trained via XGBoost \cite{gui2024conformal}.

Based on the observations from the previous experiment, we start by analyzing the calibration properties of the model. To this end, Fig.~\ref{fig:radio_rd} presents the reliability diagram \cite{guo2017calibration} for the TLDNN cloud model and the LSTM-DAE edge model. Consistent with the observations in Fig.~\ref{fig:image_rd}, the cloud model exhibits close alignment between accuracy and confidence, whereas the edge model shows a non-negligible gap between the two, indicating comparatively worse calibration.

\begin{figure} [!htb] 
    \centering
    \centerline{\includegraphics[scale=0.21]{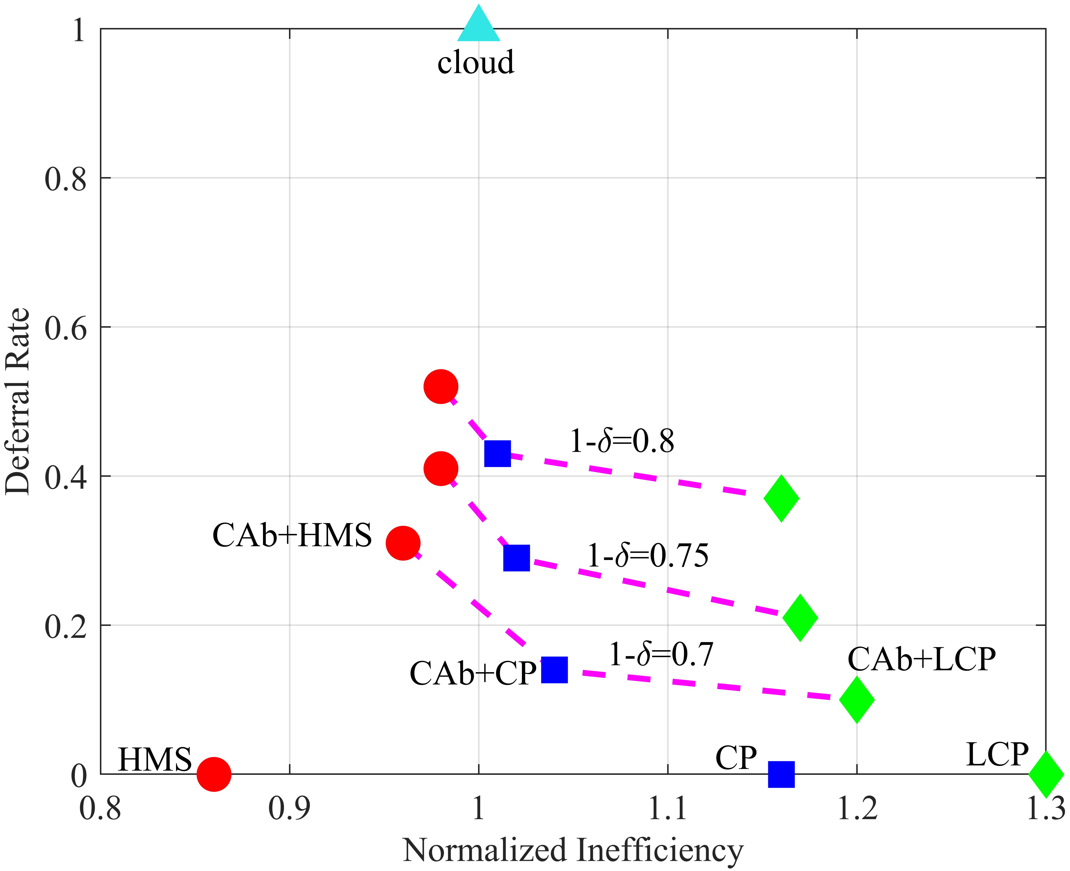}}
    \caption{Deferral rate versus normalized inefficiency obtained by changing the target average satisfaction level $1-\delta$ on the RadioML2018.01a dataset for CAb schemes, namely CAb+HMS, CAb+CP, and CAb+LCP. Points on the same dashed line share the same target average satisfaction rate $1-\delta$.}
    \label{fig:radio_pareto} 
\end{figure}

\noindent \textbf{Confidence-based versus conformal alignment-based model cascading.} Fig.~\ref{fig:radio_CA} reports average satisfaction rate, deferral rate, and normalized inefficiency for CbD and CAb schemes as a function of the target average satisfaction level in the set $1-\delta \in \{ 0.6, 0.65, \dots, 0.95\}$ with a fixed conditional coverage requirement $1-\alpha=0.8$. This figure offers conclusions aligned with Fig.~\ref{fig:image_CA}: (\emph{i}) the proposed CAb schemes consistently guarantee the condition (\ref{eq:batch_goal}), irrespective of the choice of the edge prediction set; (\emph{ii}) the CbD schemes fail to meet the target reliability requirement (\ref{eq:batch_goal}); and (\emph{iii}) a trade-offs exists between deferral rate and prediction set size, whereby a smaller prediction set results in a higher deferral rate, and conversely.

\noindent \textbf{Trade-off between deferral rate and normalized inefficiency.}  To further illustrate the discussed trade-off between the deferral rate and normalized inefficiency for the CAb schemes, Fig.~\ref{fig:radio_pareto} plots these two metrics as a function of the target average satisfaction levels $1-\delta$. The results confirm that stricter reliability requirements lead to higher deferral rates for all schemes, whereas larger prediction sets tend to correspond to lower deferral rates.

\begin{figure} [tb] 
    \centering
    \vspace{-0.5cm}
    \centerline{\includegraphics[scale=0.23]{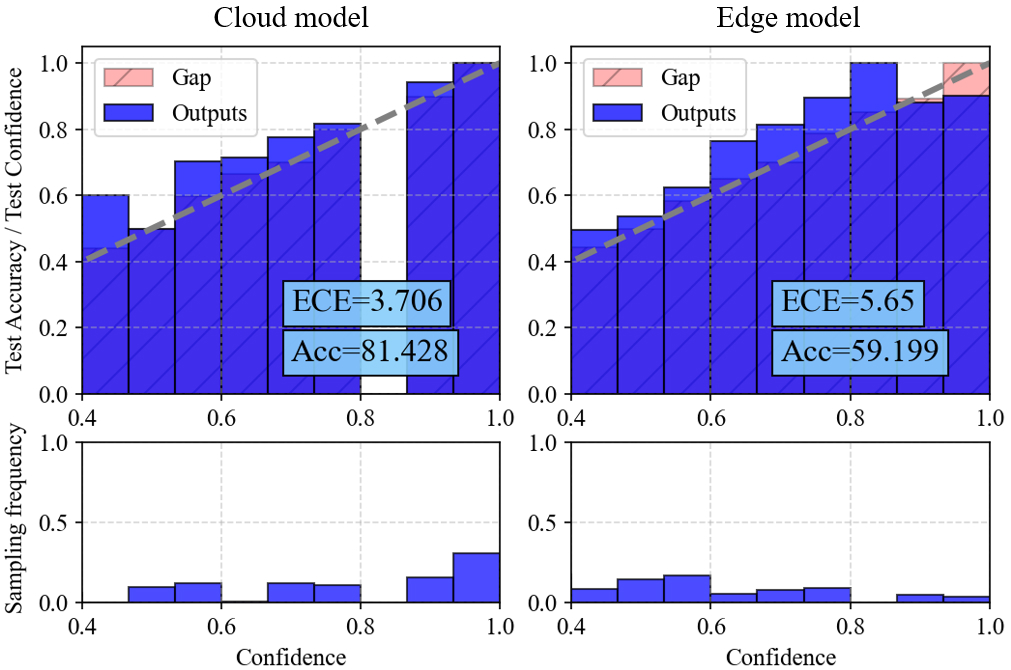}}
    \caption{Reliability diagram for the Qwen2-7B-Instruct cloud model (left) and the Qwen2-1.5B-Instruct edge model (right), evaluated on the TeleQnA dataset \cite{maatouk2023teleqna}.}
    \vspace{-0.3cm}
    \label{fig:qa_rd} 
\end{figure}

\begin{figure*} [!b] 
    \centering
    \centerline{\includegraphics[width=\textwidth]{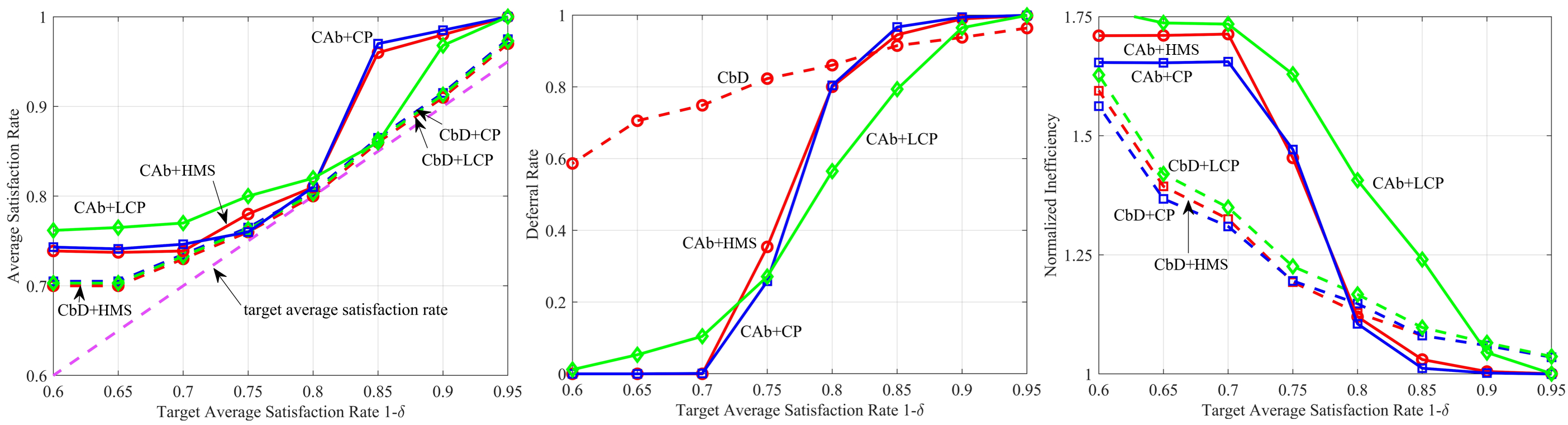}}
    \caption{Average satisfaction rate (left), deferral rate (middle), and normalized inefficiency (right) for CbD schemes and CAb schemes versus different target average satisfaction levels $1-\delta \in \{ 0.6, 0.65, \dots, 0.95\}$ on the TeleQnA dataset for required conditional coverage level $1-\alpha=0.8$. The dashed straight line in the figure indicates the target average satisfaction rate $1-\delta$. The other dashed lines in the figure represent the CbD schemes. }
    \label{fig:qa_CA} 
\end{figure*}

\subsection{Question Answering}
We now consider TeleQnA \cite{maatouk2023teleqna}, a real-world multiple-choice QA dataset, which is used for assessing the knowledge of LLMs in the field of telecommunications. The TeleQnA dataset contains $10,000$ multiple-choice questions, including $6,441$ five questions-options pairs, $3,456$ four questions-options pairs, and a small number with two or three questions-options pairs, spanning five distinct categories: lexicon, research overview, research publications, standards overview, and standards specifications. We focus here on the four questions-options pairs.

We adopt language models Qwen2-7B-Instruct and Qwen2-1.5B-Instruct \cite{qwen2} as the cloud and edge models, respectively, without fine-tuning. Treating the LLMs as black boxes, we approximate the cloud conditional distribution $p^*(y|x)$ and edge conditional distribution $p^e(y|x)$ by randomly sampling $10$ answers per question as in \cite{wang2022probabilistic}. The alignment score predictor $\hat{C}(x)$ is trained via XGBoost \cite{gui2024conformal}.

Based on the insights obtained from the previous experiment, we analyze the calibration properties of the model in Fig.~\ref{fig:qa_rd}. This figure shows the reliability diagram \cite{guo2017calibration} for the Qwen2-7B-Instruct cloud model and Qwen2-1.5B-Instruct edge model. It is observed that, in stark contrast to the previous setting, here the edge model is generally under-confident but well-calibrated, exhibiting small negative gaps between accuracy and confidence. As we will see, this modifies the relative performance of cascading schemes based on HMS, CP, and LCP prediction sets, as compared to the previous experiment.

\noindent \textbf{Confidence-based versus conformal alignment-based model cascading.} To elaborate, in a manner similar to Fig.~\ref{fig:image_CA}, Fig.~\ref{fig:qa_CA} evaluates average satisfaction rate, deferral rate, and normalized inefficiency for CbD and CAb schemes against the target average satisfaction level in the set $1-\delta \in \{ 0.6, 0.65, \dots, 0.95\}$ with a fixed conditional coverage requirement $1-\alpha=0.8$.

Since the edge model is better calibrated, even CbD schemes can meet the target average satisfaction requirement (\ref{eq:batch_goal}) in this example. It is emphasized, however, that this is a purely empirical observation, and there is a priori no guarantee that CbD schemes would satisfy the condition (\ref{eq:batch_goal}). In contrast, CAb schemes adapt the overly conservative prediction regions generated by the under-confident edge model into less conservative regions that still provably satisfy the average satisfaction rate requirement (\ref{eq:batch_goal}). Furthermore, CAb schemes are seen in the figure to obtain far lower deferral rates and only modest increases in normalized inefficiency, as compared to CbD methods. For instance, at the fixed target average satisfaction level $1-\delta=0.75$, CAb methods reduce deferral rate by approximately $60\%$, while incurring a $20\%$ increase in normalized inefficiency.

In terms of the relative performance of different prediction sets, while LCP continues to produce the largest prediction sets with the smallest deferral rate, HMS and CP exhibit similar deferral rates and normalized inefficiency levels, especially for a higher target average satisfaction requirement, e.g., $1-\delta \geq 0.8$. This result is expected given that a well-calibrated model generally yields HMS with good marginal coverage guarantees.

\begin{figure} [tb] 
    \centering
    \centerline{\includegraphics[scale=0.21]{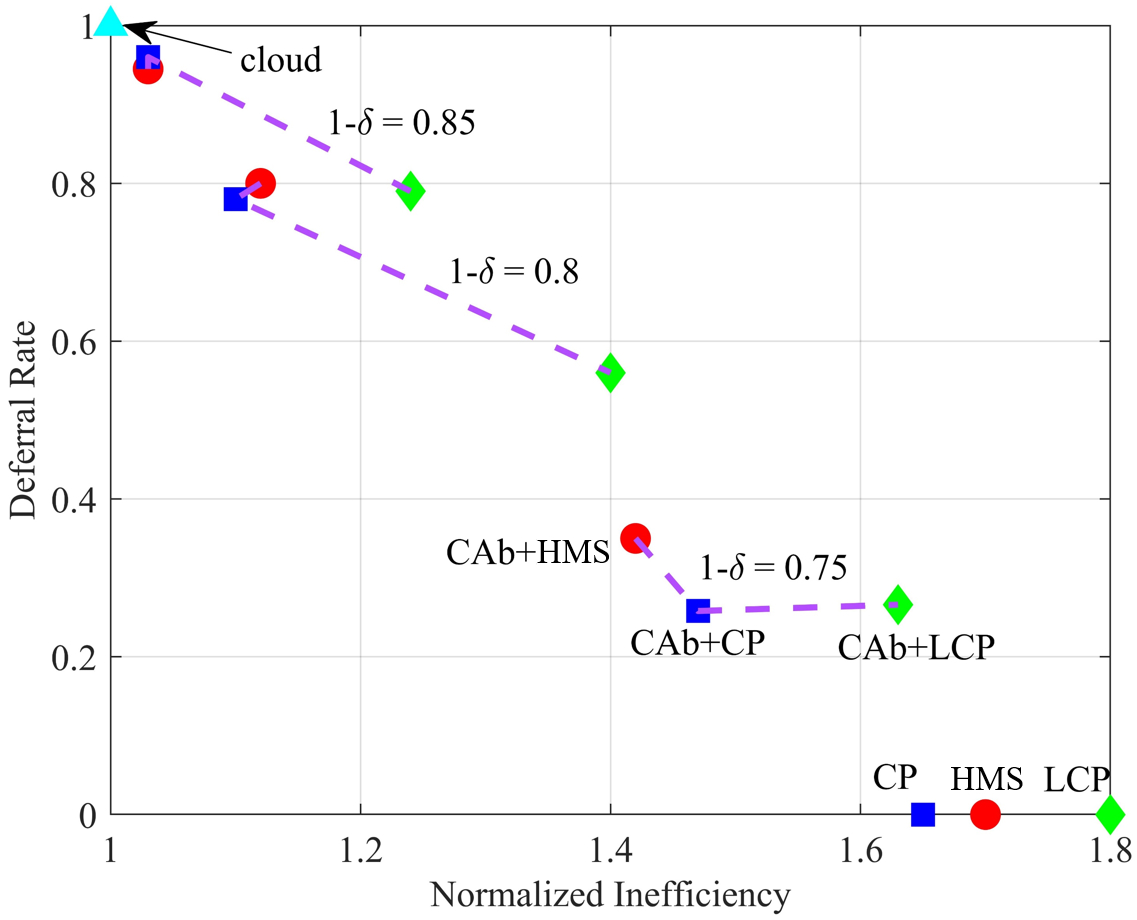}}
    \caption{Deferral rate versus normalized inefficiency obtained by changing the target average satisfaction level $1-\delta$ on TeleQnA dataset for CAb schemes, namely CAb+HMS, CAb+CP, and CAb+LCP. Points on the same dashed line share the same target average satisfaction rate $1-\delta$.}
    \label{fig:qa_pareto} 
\end{figure}

\noindent \textbf{Trade-off between deferral rate and normalized inefficiency.} To further elaborate on this point, Fig.~\ref{fig:qa_pareto} demonstrates the trade-off between the deferral rate and normalized inefficiency for CAb schemes by varying the target average satisfaction levels $1-\delta$. The figure confirms that with a better calibrated model, CP and HMS tend to yield similar results in both deferral rate and normalized inefficiency, while LCP remains the most conservative solution, producing the largest prediction sets with the lowest deferral rate.

\section{Conclusion} \label{sec:conclusion} 
In this paper, we have proposed a novel edge-cloud model cascading mechanism producing prediction sets that have the same conditional coverage properties of sets produced at the cloud model only. The proposed method, namely conformal alignment-based (CAb) model cascading, provides statistical guarantees on the average fraction of edge-processed decisions that satisfy cloud-level conditional coverage, while minimizing reliance on cloud resources. This guarantee is achieved by casting the escalation from edge to cloud models as a multiple-hypothesis testing (MHT) problem, where the conditional coverage probability serves as the tailored alignment score. Empirical results demonstrate that CAb methods exhibit a tunable trade-off among conditional coverage, deferral rate, and set size. For instance, compared to confidence-based deferral (CbD) schemes, at the fixed target average satisfaction rate $1-\delta=0.75$, CAb schemes reduce deferral rate by approximately $60\%$, at the cost of $20\%$ increase in set size, while maintaining provable reliability guarantees.

Future research directions may include evaluating the robustness of CAb schemes under covariate shift \cite{tibshirani2019conformal}, extending the proposed framework to localized conformal alignment \cite{wu2024conditional}, and integrating conformal e-values into the alignment process to offer anytime-valid guarantees for sequential inputs in edge-cloud systems \cite{ramdas2024hypothesis, xu2021unified}.

\bibliographystyle{IEEEtran}
\bibliography{refs}
\newpage
\clearpage

\title{Appendix for Reliable Inference in Edge-Cloud Model Cascades via Conformal Alignment \\
}

\author{Jiayi Huang, Sangwoo Park,~\IEEEmembership{Member,~IEEE}, Nicola Paoletti,  and Osvaldo Simeone,~\IEEEmembership{Fellow,~IEEE}}

\maketitle

{\appendix[Proof of Proposition.~1]

Given the output $\mathcal{S}= \mathcal{D}_{(\ell_{\text{CA}})}^{\text{uns,te}}$ of the CAb model cascading methodology, by the definition of the FDR in (23), we have

\begin{align}  \label{eq:appendix_proof}
   & \text{FDR}  =  \mathbb{E} \left[ 
\frac{\left| \{x_i \in \mathcal{D}^{\text{uns,te}}_{( \ell_{\text{CA}})}: C^*(x_i) < 1-\alpha      \} \right|}
{\left|\mathcal{D}^{\text{uns,te}}_{( \ell_{\text{CA}})}\right|} \right] \nonumber  \\ & \overset{(a)}{=} \mathbb{E} \left[ \widehat{\text{FDP}}_{( \ell_{\text{CA}})} \cdot \frac{1 + |\mathcal{D}^{\text{val}}|}{|\mathcal{D}^{\text{te}}|}  \frac{\left| \{x_i \in \mathcal{D}^{\text{uns,te}}_{( \ell_{\text{CA}})}: C^*(x_i) < 1-\alpha      \}\right|}{1+\left| \{x_i \in \mathcal{D}^{\text{uns,val}}_{( \ell_{\text{CA}})}: C^*(x_i) < 1-\alpha      \}\right|}   \right] \nonumber \\
& \overset{(b)}{\leq} \delta \cdot \frac{1 + |\mathcal{D}^{\text{val}}|}{|\mathcal{D}^{\text{te}}|} \mathbb{E} \left[ \frac{\left| \{x_i \in \mathcal{D}^{\text{uns,te}}_{( \ell_{\text{CA}})}: C^*(x_i) < 1-\alpha      \}\right|}{1+\left| \{x_i \in \mathcal{D}^{\text{uns,val}}_{( \ell_{\text{CA}})}: C^*(x_i) < 1-\alpha\}\right|}  \right], \tag{32}
\end{align}
where $(a)$ follows from the definition of the FDP estimator in (30), and $(b)$ is obtained by adopting the stopping rule in (31), which implies the inequality $\mathbb{E} \left[\widehat{\text{FDP}}_{(\ell_{\text{CA}})}\right] \leq \delta$.

For notational convenience, we define the data-dependent statistic
\begin{align} \label{eq:martingale_RV}
    \mathcal{M}_{(\ell)} = \frac{\left| \{x_i \in \mathcal{D}^{\text{uns,te}}_{( \ell)}: C^*(x_i) < 1-\alpha      \}\right|}{1+\left| \{x_i \in \mathcal{D}^{\text{uns,val}}_{( \ell)}: C^*(x_i) < 1-\alpha\}\right|}. \tag{33}
\end{align}
The sequence $\{ \mathcal{M}_{(\ell)}\}_{\ell \geq 0}$ can be shown to be a super-martingale with respect to the filtration $\{ \mathcal{F}_{(\ell)}\}_{\ell \geq 0}$, where
\begin{align}
    \mathcal{F}_{(\ell)}  = \sigma ( \{x_i, C^*(x_i), \hat{C}(x_i)\}_{i=1}^{\ell} ) \tag{34}
\end{align}
represents the information set observed up to screening step $\ell$ \cite[Lemma~1]{gui2025acs}. Therefore, by the defining condition of a super-martingale, we have the inequality
\begin{align} \label{eq:martingale_definition}
    \mathbb{E} \left[\mathcal{M}_{(\ell+1)} |\mathcal{F}_{(\ell)}\right] \leq \mathcal{M}_{(\ell)}. \tag{35}
\end{align}
Then, taking expectations on both sides in (\ref{eq:martingale_definition}) and applying the law of iterated expectations, we have the inequality
\begin{align} \label{eq:martingale_property_1}
    \mathbb{E} \left[\mathcal{M}_{(\ell+1)}\right] \leq \mathbb{E} \left[ \mathcal{M}_{(\ell)} \right]. \tag{36}
\end{align}

Furthermore, since the sequential screening process terminates at step $\ell_{\text{CA}}$, we can leverage the optional stopping theorem for super-martingale \cite{billingsley2017probability} to obtain the inequality 
\begin{align} \label{eq:martingale_property_2}
    \mathbb{E} \left[\mathcal{M}_{(\ell_{\text{CA}})}\right] \leq \mathbb{E} \left[ \mathcal{M}_{(0)} \right], \tag{37}
\end{align}
where, by the definition of $\mathcal{M}_{(\ell)}$ in (\ref{eq:martingale_RV}), we have 
\begin{align} \label{eq:martingale_bounds}
    \mathbb{E} \left[ \mathcal{M}_{(0)} \right] &= \mathbb{E} \left[ \frac{\left| \{x_i \in \mathcal{D}^{\text{te}}: C^*(x_i) < 1-\alpha      \}\right|}{1+\left| \{x_i \in \mathcal{D}^{\text{val}}: C^*(x_i) < 1-\alpha\}\right|} \right] \nonumber \\ &=  \frac{|\mathcal{D}^{\text{te}}|}{1 + |\mathcal{D}^{\text{val}}|}. \tag{38}
\end{align}

Finally, using (\ref{eq:appendix_proof}), (\ref{eq:martingale_property_2}), and (\ref{eq:martingale_bounds}), we obtain the desired result
\begin{align}
    \mathbb{E}\bigg[\frac{ | \{x_i \in \mathcal{S}: \Pr[y_i \in \Gamma(x_i)|x_i] \geq 1-\alpha \} | }{|\mathcal{S}|}\bigg] &= 1- \text{FDR} \nonumber \\ &\geq 1 - \delta, \tag{39}
\end{align}
which ensures that the output $\mathcal{S}= \mathcal{D}_{(\ell_{\text{CA}})}^{\text{uns,te}}$ of the CAb model cascading methodology satisfies the guarantee (6).
}

\end{document}